\documentclass[twoside]{article} 
\usepackage{PRIMEarxiv}

\usepackage{inputenc}
\usepackage[T1]{fontenc}    
\usepackage{hyperref}       
\usepackage{url}            
\usepackage{booktabs}       
\usepackage{amsfonts}       
\usepackage{nicefrac}       
\usepackage{lipsum}
\usepackage{fancyhdr}       
\usepackage{graphicx}       
\graphicspath{{media/}}     

\usepackage{float}

\usepackage{multicol,lipsum}
\usepackage{enumerate}

\usepackage[table, dvipsnames]{xcolor}

\renewcommand{\headrulewidth}{.4pt}

\usepackage{graphicx}
\usepackage{afterpage}
\usepackage{graphicx}
\usepackage{amssymb}
\usepackage{amsmath}
\usepackage{bm}
\usepackage{hyperref}
\usepackage{float, multirow}
\usepackage{xcolor}
\usepackage{setspace}
\usepackage{mathptmx} 
\usepackage{newtxtext,newtxmath}
\usepackage{lscape}
\usepackage{soul} 
\usepackage[title]{appendix}
\usepackage{setspace}
\usepackage{hanging}
\usepackage{xspace}
\usepackage{tikz}
\usetikzlibrary{positioning}
\usetikzlibrary{shapes.geometric, arrows}
\usepackage{calc}
\usetikzlibrary{arrows.meta}
\usepackage{varwidth}
\usepackage[numbers, sort&compress, square]{natbib}
\usepackage{hyperref}
\usepackage{cleveref}
\hypersetup{colorlinks,linkcolor={blue},citecolor={blue},urlcolor={blue}}
\usepackage{tabulary}
\newcolumntype{L}[1]{>{\raggedright\let\newline\\\arraybackslash\hspace{0pt}}m{#1}}
\newcolumntype{C}[1]{>{\centering\let\newline\\\arraybackslash\hspace{0pt}}m{#1}}
\newcolumntype{R}[1]{>{\raggedleft\let\newline\\\arraybackslash\hspace{0pt}}m{#1}}

\usepackage{subcaption}
\makeatletter
\renewcommand{\fnum@figure}{Fig. \thefigure}									 
\makeatother

\usepackage[singlelinecheck=off, skip=2pt,font=small]{caption}
\usepackage{lipsum}

\newlength\Myfigwd

\Crefname{figure}{Fig.}{Fig.}
\Crefname{fig_a}{Fig.}{Fig.}
\creflabelformat{Fig_a}{#2{\color{NavyBlue}#1a)}#3}
\Crefname{fig_b}{Fig.}{Fig.}
\creflabelformat{Fig_b}{#2{\color{NavyBlue}#1b)}#3}
\Crefname{fig_c}{Fig.}{Fig.}
\creflabelformat{Fig_c}{#2{\color{NavyBlue}#1c)}#3}
\Crefname{equation}{Eq.}{Eq.}
\Crefname{section}{Section}{Section}

\usepackage{enumitem}

\usepackage{fancyhdr}
\usepackage{lineno}
\usepackage{stackengine}

\newlist{romanitem}{enumerate}{1}
\setlist[romanitem,1]{label=\textbf{Stage \Roman*}:, align=left, leftmargin=*}

\usepackage{graphicx,wrapfig,lipsum}

\usepackage{anyfontsize}
\title{Leveraging Deep Operator Networks $\bigl($DeepONet$\bigr)$ for Acoustic Full Waveform Inversion $\bigl($FWI$\bigr)$}

\author{
    Kamaljyoti Nath\\
    Division of Applied Mathematics,\\
    Brown University, \\
    Providence, RI, USA\\
    kamaljyoti\_nath@brwon.edu
    \And
    Khemraj Shukla\\
    Division of Applied Mathematics,\\
    Brown University, \\
    Providence, RI, USA\\
    khemraj\_shukla@brown.edu
    \And
    Victor C. Tsai  \textsuperscript{*}\\
    Dept. of Earth, Environmental and Planetary Sciences,\\
    Brown University, \\
    Providence, RI, USA\\
    victor\_tsai@brown.edu
    \And
    Umair bin Waheed\\
    Dept. of Geosciences, \\
    King Fahd University of Petroleum and Minerals, \\
    Dhahran, Eastern Province, Saudi Arabia\\
    umairbin.waheed@kaust.edu.sa
    \And 
    Christian Huber\\
    Dept. of Earth, Environmental and Planetary Sciences,\\
    Brown University, \\
    Providence, RI, USA\\
    christian\_huber@brown.edu
    \And
    Omer Alpak\\
    Shell USA, Houston\\
    Omer.Alpak@shell.com
    \And 
    Chuen-Song Chen\\
    Shell USA, Houston\\
    Chuen-Song.Chen@shell.com
    \And
    Ligang Lu\\
    Shell USA, Houston\\
    Ligang.Lu@shell.com
    \And
    Amik St-Cyr \\
    Shell USA, Houston\\
    Amik.St-Cyr@shell.com
}
\begin{document}
\maketitle
\fancypagestyle{alim}{\fancyhf{}\renewcommand{\headrulewidth}{0pt}\fancyfoot[R]{\today}\fancyfoot[L]{Preprint}}
\thispagestyle{alim}
\vspace{-0.5cm}
\begin{abstract}
Full Waveform Inversion (FWI) is an important geophysical technique considered in subsurface property prediction. It solves the inverse problem of predicting high-resolution Earth interior models from seismic data. Traditional FWI methods are computationally demanding. Inverse problems in geophysics often face challenges of non-uniqueness due to limited data, as data are often collected only on the surface. In this study, we introduce a novel methodology that leverages Deep Operator Networks (DeepONet) to attempt to improve both the efficiency and accuracy of FWI. The proposed DeepONet methodology inverts seismic waveforms for the subsurface velocity field. This approach is able to capture some key features of the subsurface velocity field. We have shown that the architecture can be applied to noisy seismic data with an accuracy that is better than some other machine learning methods. We also test our proposed method with out-of-distribution prediction for different velocity models. The proposed DeepONet shows comparable and better accuracy in some velocity models than some other machine learning methods. To improve the FWI workflow, we propose using the DeepONet output as a starting model for conventional FWI  and that it may improve FWI performance. While we have only shown that DeepONet facilitates faster convergence than starting with a homogeneous velocity field, it may have some benefits compared to other approaches to constructing starting models. This integration of DeepONet into FWI may accelerate the inversion process and may also enhance its robustness and reliability.
\end{abstract}
\let\thefootnote\relax\footnotetext{\textsuperscript{*} Corresponding author: Victor C. Tsai: (victor\_tsai@brown.edu)}
\section{Introduction}
\label{Section:Introduction}
Understanding the Earth's interior structure is fundamental to advancements in reservoir characterization, monitoring of the movement of subsurface fluids for energy recovery and energy transition applications such as CO\textsubscript{2} storage, seismic hazard assessment, and geotechnical applications~\citep{nolet2012seismic}. Full waveform inversion (FWI) has been widely adopted for reconstructing detailed subsurface velocity models because it fully exploits seismic wavefield data, including amplitude and phase information~\citep{tarantola1984inversion,Fichtner_2010_book}. By leveraging the sensitivity of seismic wavefields to variations in subsurface properties, FWI can resolve complex geological features with high resolution and accuracy. This process typically involves modeling the propagation of seismic waves through the Earth and iteratively updating the velocity model to minimize the misfit between observed and simulated data. Unlike traditional seismic imaging techniques, which rely primarily on travel times or reflected wave amplitudes, FWI incorporates the entire wavefield, enabling it to capture subtle variations in subsurface properties and delineate fine-scale structures critical for both scientific and industrial applications~\citep{Virieux_2009}.
\par Over the years, numerous approaches have been developed to improve the performance of FWI, driven by the demand for high-resolution subsurface imaging and the need to address its limitations. These efforts have focused on refining optimization techniques, improving data preprocessing, and developing strategies to better handle complex wave phenomena~\citep{pratt1998gauss, fichtner2013multiscale, metivier2013full, alkhalifah2012taming, warner2016adaptive}. Despite these advancements, FWI remains computationally expensive and faces significant obstacles in practical applications. One of the major difficulties lies in its sensitivity to the initial velocity model. An inaccurate starting model can cause the optimization process to converge toward incorrect solutions. Another persistent challenge is the lack of low-frequency data in seismic recordings, which limits FWI's ability to recover large-scale velocity variations and often leads to cycle-skipping issues. Additionally, the computational cost of modeling wave propagation through complex subsurface structures is substantial, particularly in three-dimensional settings. Factors such as data noise, sparse receiver coverage, and the need for high-fidelity forward simulations further complicate the process. These challenges underscore the ongoing need for innovative approaches to make FWI more robust and computationally feasible.
\par With advancements in neural network architectures, researchers have increasingly focused on neural network-based methods due to their ability to learn mappings between high-dimensional data spaces and functions. A particularly promising approach within this domain is operator learning \cite{Chen_1995_universal_operator}, which utilizes neural networks as parameterized high-dimensional functions to establish mappings between two functional spaces. For instance, consider learning the derivative of a function \( f(x) \), expressed as \( \frac{\mathrm{d} f}{\mathrm{d}x} = g(x) \). The goal is to approximate the derivative operator \( \frac{\mathrm{d}}{\mathrm{d}x} \) using a neural network. The input to this neural network comprises a set of admissible functions for differentiation, along with the domain of \( f(x) \), denoted as \( x \). The neural network outputs \( g(x) \), the derivative of \( f(x) \). Since both \( f(x) \) and \( g(x) \) are functions, the mapping must operate between two functional spaces, making the operator a mapping between these spaces. In this study, the governing partial differential equation for wave propagation is a second-order PDE given by \(\frac{\partial^2 u}{\partial t^2} = c^2 \Delta u = 0\), subject to specified initial and boundary conditions. One of the primary objectives is to learn the operator \((\partial_t^2 - c^2 \Delta)\) for a given spatially varying wave speed \(c(x, y)\), ensuring that \((\partial_t^2 - c^2 \Delta)u = 0\). This is commonly referred to as the forward problem in seismic modeling. In the inverse problem, the aim is to approximate \(c(x, y)\) when \(u\) is provided, such that the relationship \((\partial_t^2 - c^2 \Delta)u = 0\) holds. In this work, we propose a neural operator that learns the operator \((\partial_t^2 - c^2 \Delta)\) for a given \(u(x, t)\), ensuring the condition \((\partial_t^2 - c^2 \Delta)u = 0\) is satisfied.

\par In this study, we develop the Deep Operator Network (DeepONet) based architecture proposed by \citeauthor{Lu_2021_DeepONet} \citep{Lu_2021_DeepONet}. DeepONet is a neural network based operator approximation method, which maps the space of input functions to the space of output functions and is based on the universal approximation theorem for operators proposed by \citeauthor{Chen_1995_universal_operator} \cite{Chen_1995_universal_operator}. In DeepONet, the one-hidden layer networks proposed in \cite{Chen_1995_universal_operator} are replaced by two deep neural networks: branch and trunk networks. The branch network processes the input function, capturing its functional characteristics, while the trunk network handles the domain coordinates where the output is evaluated. Together, these networks effectively encode and reconstruct the operator mapping, enabling DeepONet to generalize across infinite-dimensional spaces with high accuracy and efficiency.
\par One of the major advantages of DeepONet is that it can be applied to simulation data, experimental data or both \citep{Higgins_2021}. There are also no restrictions on the type of network considered in the branch and trunk. Furthermore, DeepONet can predict the output not only on a uniform grid but also at arbitrary points in the domain. DeepONet was further developed for different flavours of problems, with POD-DeepONet \cite{Lu_2022_Comp_deep_fno}, MIONet \cite{Pengzhan_2022_MIONet}, Causality DeepONet \cite{Liu_2022_Causality}, Fourier-DeepONet \cite{Zhu_2023_fourier}, En-DeepONet \cite{Ehsan_2024} a few variations of the original DeepONet structure. Other researchers are also focusing on developing other neural operators, including Fourier Neural Operators (FNO) \cite{Li_2020_FNO}, which leverage Fourier transforms, graph kernel networks \cite{Li_2020_GKN}, Wavelet Neural Operators (WNO) \cite{Tripura_2023_WNO}, and Laplace neural operators \cite{cao2024laplace}. Researchers have also studied DeepONet with Fourier layers \cite{Zhu_2023_fourier} in their neural network architecture. A comprehensive and FAIR comparison between DeepONet and FNO can be found on \cite{Lu_2022_Comp_deep_fno}. 
\par Neural network-based methods have been considered in FWI and other geotechnical problems in the past. For example, \citeauthor{Majid_2022_FWI} \cite{Majid_2022_FWI} studied FWI using physics-informed neural network. InversionNet, which considers convolutional and deconvolutional neural networks, was considered by \citeauthor{Yue_2020_InversionNet} \cite{Yue_2020_InversionNet} and \citeauthor{OpenFWI} \cite{OpenFWI}, where images of observed wavefields are mapped to the image of the subsurface velocity field. \citeauthor{Zhongping_2020_VelocityGAN} \cite{Zhongping_2020_VelocityGAN} and \citeauthor{OpenFWI} \cite{OpenFWI} studied FWI using velocityGAN. \citeauthor{Peng_2021_UPFWI} \cite{Peng_2021_UPFWI}, \citeauthor{OpenFWI} \cite{OpenFWI} considered UPFWI for the inverse problem of FWI, which is an unsupervised method where the forward model described by the differential equations are considered in the training loop. \citeauthor{OpenFWI} \cite{OpenFWI} also made a comparison study of the three methods (InversionNet, VelocityGAN and UPFWI) for different types of velocity models. \citeauthor{Taccari_2024_Ground_water} \cite{Taccari_2024_Ground_water} consider DeepONet for groundwater modelling, \citeauthor{Ehsan_2024} \cite{Ehsan_2024} proposed Enriched-DeepONet (En-DeepONet) for Eikonal operator, \citeauthor{Zhu_2023_fourier} \cite{Zhu_2023_fourier} considered a hybrid method Fourier-DeepONet where the authors considered varying source parameters for FWI. \citeauthor{Lehmann_2024_FFNO} \cite{Lehmann_2024_FFNO} studied 3D wave elastic wave propagation using Factorized Fourier Neural Operator (F-FNO), \citeauthor{Waleed_2024_U-DeepONet} \cite{Waleed_2024_U-DeepONet} studied geologic carbon sequestration using a U-Net enhanced DeepONet. \citeauthor{Alpak_2023_ADR} \cite{Alpak_2023_ADR} proposed FNO-based deep residual network (FResNet++) for Two-Phase Flow and Transport Simulations.
\par In the present study, we will formulate a neural network-based operator for reconstructing the subsurface velocity model from acoustic data. Specifically, we will leverage the Deep Operator Network (DeepONet) to develop a neural operator for the inverse problem. The proposed DeepONet architecture predicts the velocity field. In the next section (\Cref{Subsection:Problem statement}), we will discuss the problem setup considered in the present study and the governing equation for acoustic waves in isotropic media.
\par The rest of the article is organized as follows:  \Cref{Section:Methodology} discusses the methodology implemented in the present study, which is DeepONet (\Cref{Subsection:DeepONet}) and proposed workflow pairing DeepONet and FWI method where a-priori velocity field is optained from DeepONet to perform acoustic FWI loop (\Cref{Subsection:Lagecy FWI}) and data generation (\Cref{Subsection:Data generation}); in the following section, \Cref{Section:Numerical results: DeepONet}, we will validate our proposed DeepONet with numerical results for different input conditions and comparison with InversionNet; the conclusion of the present study are presented in  \Cref{Section:Summary}.

\subsection{Problem statement}
\label{Subsection:Problem statement}
Seismic waves propagate through the subsurface and are reflected at interfaces between different geological layers, providing critical information about the Earth's interior. Seismic waves, generated using controlled sources such as vibrators or explosives, are recorded at the surface using seismic sensors like geophones. By analyzing these reflected waves, it is possible to reconstruct high-resolution images of the subsurface using FWI. Mathematically, the propagation of seismic waves is described by the wave equation. The prediction of seismic waveforms for a given subsurface velocity model and the source is termed the forward problem, whereas determining the subsurface velocity field from observed seismic data constitutes the inverse problem. A key challenge in solving the inverse problem lies in the limited availability of data, as seismic recordings are typically confined to the surface, leaving deeper regions poorly illuminated \cite{dellinger2024forensic}. Despite these limitations, FWI remains a powerful tool for subsurface imaging, leveraging the complete seismic wavefield to achieve detailed and accurate velocity models.

\par
\begin{figure}[H]
    \centering
    \includegraphics[width=0.75\linewidth]{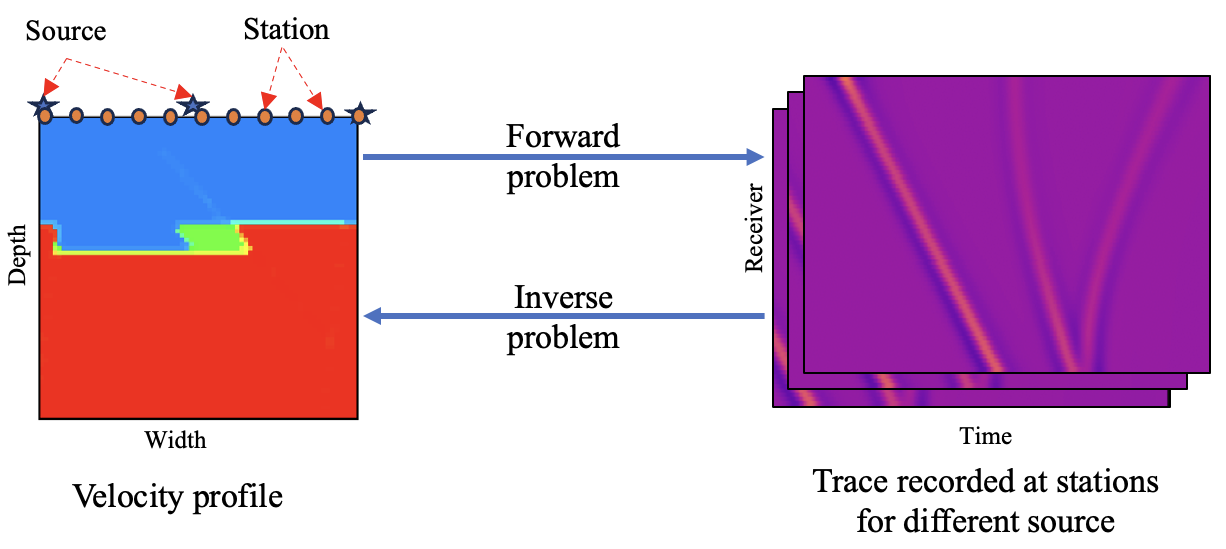}
    \caption{\textbf{A schematic representation of the problem setup} illustrating the forward problem (predicting seismic waveforms at receivers for a given subsurface velocity model and source) and the inverse problem (estimating the subsurface velocity field from recorded seismic waveforms). The forward problem models the data generation process, while the inverse problem focuses on reconstructing the velocity profile from the recorded seismic data. In the present study, our focus is the inverse problem of velocity field prediction using recorded seismic waveforms.}
    \label{Fig:Problem setup}
\end{figure}

 \Cref{Fig:Problem setup} presents a schematic diagram of the forward and inverse problem in the context of a 2D velocity field. Seismic waves generated by the sources travel through the subsurface, where they interact with geological layers through reflection and refraction. They are recorded at seismic stations typically located at the surface. Each source is excited independently, and the waves are recorded for each source. It is generally assumed that the recorded wavefield is free from interference caused by other sources, although noise is often present in the recorded seismic data \cite{Yilmaz_2001_seismic}. The equation governing acoustic waves traveling through an isotropic media can be written as,
\begin{equation}
    \nabla^2u(x, y, t) - \dfrac{1}{c^2(x,y)}\dfrac{\partial^2 u(x,y,t)}{\partial t^2} = s(t),\;\;\;\;\;\; \nabla^2=\dfrac{\partial^2}{\partial x^2} + \dfrac{\partial^2}{\partial y^2},
    \label{Eq:Governing equation}
\end{equation}
where $c(x,y)$ is the velocity field of the subsurface, $u(x,y,t)$ is the pressure field as a function of space and time. In the case of the forward problem, the the velocity field ($c(x,y)$) are known, and the objective is to evaluate the pressure field $u(x,y,t)$. Conversely, in the inverse problem, the objective is to predict the velocity field ($c(x,y)$) of the subsurface when the pressure field $u(x,y,t)$. However, in most inverse problems, observations of the state variables are limited to specific locations. Typically, for practical applications, the pressure field can only be recorded at the surface, making the inverse problem inherently challenging.
\par As discussed earlier, FWI plays a crucial role in various engineering and scientific applications. In this study, our objective is to develop a robust DeepONet-based neural operator for the inverse problem of predicting the velocity field from seismic data recorded at surface stations. We assume that the source functions remain consistent across different velocity fields and that both the position of sources and stations are spatially fixed for all the input samples. Thus, mathematically, our goal is to learn an operator, $\mathcal{G}_{\bm{\theta}}$, parameterized by $\bm{\theta}$, which maps the observed pressure field at the surface, $u(x,y_0, t)$, to the corresponding velocity field, $c(x,y)$:
\begin{equation}
    \mathcal{G}_{\bm{\theta}}:u(x,y_0, t)(x,y)\rightarrow c(u(x,y_0, t))(x,y).
\end{equation}
\par Two of the major challenges in FWI are the presence of noise in the recorded seismic data, often caused by interference from other seismic sources and the practical difficulties of sensor placement in the field due to physical or logistical constraints. These limitations result in incomplete and noisy datasets, making it even more challenging to accurately predict the subsurface velocity field. In this study, we first validate the proposed DeepONet architecture using clean, simulated data. Subsequently, we evaluate its performance under more realistic conditions by introducing noise to the simulated data and testing it on datasets with both noisy and missing measurements. The missing measurement condition refers to when we do not have pressure field measurements at some of the stations, leading to zero values for these measurements.
\par Conventional FWI is highly sensitive to the initial velocity model, often leading to inaccurate solutions when the starting model is poorly defined. While neural operators, such as DeepONet, have shown promise in addressing inverse problems by learning mappings between input and output functions, they have limited extrapolation capabilities and may produce suboptimal results when applied to scenarios outside the distribution of their training data. To overcome this limitation, we propose a hybrid methodology that integrates DeepONet with conventional FWI. The velocity field predicted by DeepONet, while not fully accurate in out-of-distribution cases, can still serve as a robust initial model for FWI. This hybrid approach leverages the ability of DeepONet to provide an informed starting point and the iterative refinement capabilities of FWI to improve accuracy. To validate the proposed method, we evaluate the performance of our surrogate model on an unseen velocity field sampled from a different family of velocity models. The predictions are then used as the initial model for iterative conventional FWI. This demonstrates the effectiveness of this hybrid approach for enhanced subsurface imaging.

\section{Methodology}
\label{Section:Methodology}
\subsection{Deep Operator Networks (DeepONet)}
\label{Subsection:DeepONet}
As discussed in the previous section, the goal of this study is to develop an operator that maps the functions of seismic waveforms recorded at the surface to the corresponding velocity fields. For this purpose, we employ DeepONet as our choice of the operator learning paradigm. \citeauthor{Lu_2021_DeepONet} \citep{Lu_2021_DeepONet} proposed DeepONet based on the universal approximation theorem for operators \citep{Chen_1995_universal_operator}. It is designed to map an infinite-dimensional input function ($u\in \bm{U}$) to an output function ($v\in \bm{V}$) in a Banach space. A conventional DeepONet architecture consists of two networks: the branch network, which processes the input function sampled at fixed points, and the trunk network, which takes spatio-temporal coordinates where the output is to be evaluated. Mathematically, DeepONet can be written as $\mathcal{G}_{\bm{\theta}}:\bm{U}\rightarrow\bm{V}$ and the output is given as:
\begin{equation}
    v(\xi)\approx \mathcal{G}_{\bm{\theta}}(u)(\xi) = \sum_{i=1}^p br_i(u)tr_i(\xi),
    \label{Eq:DeepONet}
\end{equation}
where $p$ is the number of neurons in the last layer of each trunk and branch network, $br_i(u)$ and $tr_i(\xi)_i$ are the output of the branch and the trunk network, respectively. $u$ is the input function sampled at fixed $n$ points.  The DeepONet architecture is designed to be highly flexible and does not impose any specific restrictions on the type of neural network used for either the branch or trunk components. This flexibility allows adapting the architecture to suit the characteristics of the problem at hand, such as using fully connected networks, convolutional neural networks, or other specialized architectures depending on the nature of the input and output spaces. 
\par In the context of this study on FWI, we develop a DeepONet-based operator to predict the subsurface velocity field ($c(x,y)$). The branch network's input consists of seismic traces recorded at surface stations, while the trunk network takes the spatial coordinates, ($x, y$), where the velocity field needs to be evaluated. The size of the input to the branch is $n_{st}\times n_t\times n_s$, where $n_t$ is the number of points in the seismic trace, $n_{st}$ is the number of stations, and $n_s$ is the number of the seismic sources. Using a fully connected network (FCN) as the branch network is impractical as the number of neurons in the first layer (input layer) needs to be equal to $n_{st}\times n_t\times n_s$. To address this, we employ convolutional neural networks (CNNs) in the branch network. Specifically, we use two-dimensional CNNs where the input is treated as an image of size $n_{st}\times n_{t}$, with the number of sources, $n_s$, represented as channels. The branch network is thus formulated as multiple CNN layers followed by a FCN. The CNN layers extract features from the input data, while the FCN maps these features to coefficients of the basis functions defined by the trunk network.
\par
\begin{figure}[H]
    \centering
    \includegraphics[width=0.9\textwidth]{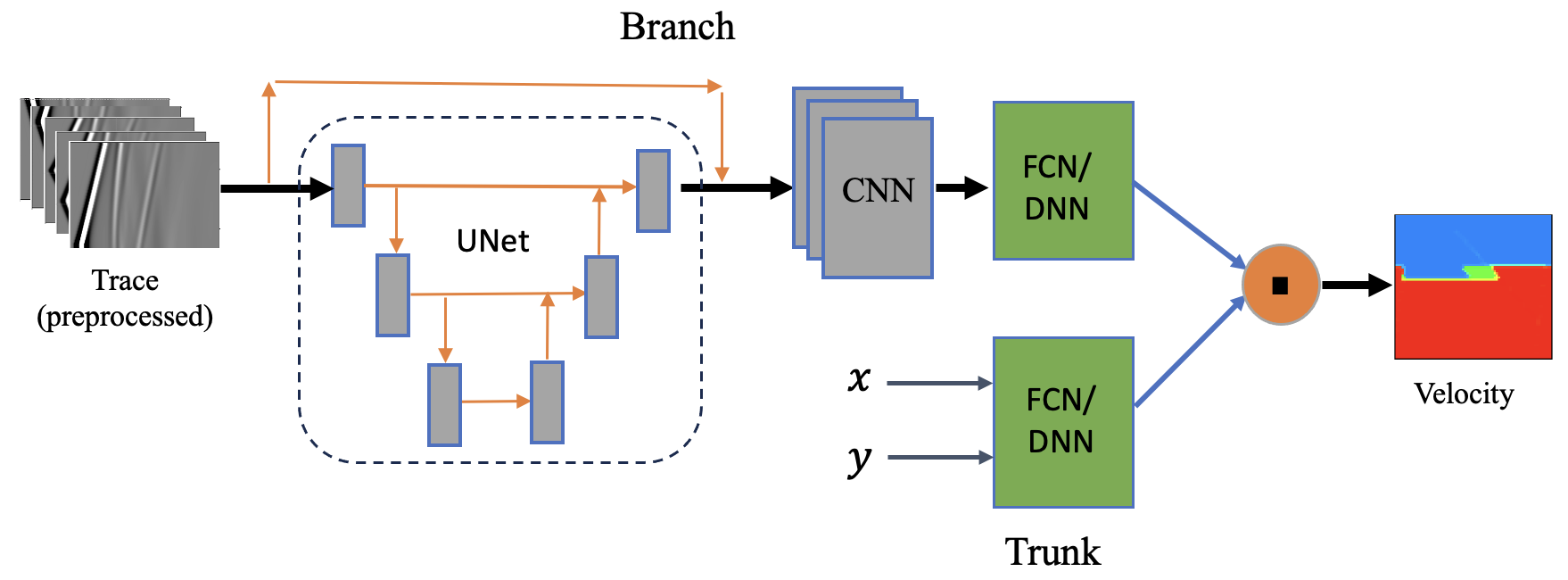}
    \caption{\textbf{Schematic diagram of the proposed DeepONet architecture:} considered in the present inverse problem. The trunk network consists of a DNN (Fully Connected feed-forward Network) with two input neurons, which take the coordinate $(x, y)$ as input where the output velocity field needs to be predicted. The branch network consists of a UNet block followed by multiple CNN layers and a DNN (Fully Connected feed-forward Network). We also added a skip connection for the input of the branch after the UNet (concatenated with the output of UNet in the channel). The input to the branch network is the pre-processed (e.g. gain function) seismic waveforms of full-waveform acoustic data for all sources recorded at stations located at the surface. The size of the input is $n_\text{station}\times n_\text{time}\times n_\text{source}$, where $n_\text{source}$ is the number of channels in the UNet / CNN. The predicted velocity field is given by the dot product of the output of the branch and trunk as shown in \Cref{Eq:DeepONet}.}
    \label{Fig:UNet-DeepONet}
\end{figure}

 To enhance multi-scale feature learning, we integrate a UNet architecture \citep{Olaf_UNet} into the branch network. A typical UNet block consists of convolution and transpose convolution layers, enabling the network to capture hierarchical spatial features effectively. In the UNet, the input is passed through a sequence of convolutional layers, followed by transpose convolutional layers. Residual connections are established between the convolutional and transpose convolutional layers by concatenating their outputs along the channel dimension. This ensures that the spatial dimensions of the input and output of the UNet remain consistent while allowing the network to leverage both fine-grained and coarse-scale features. The proposed DeepONet architecture, including the integration of the UNet into the branch network, is schematically illustrated in~\Cref{Fig:UNet-DeepONet}. This design combines the feature extraction capabilities of CNNs with the multi-scale learning advantages of UNet, making it well-suited for the complex task of predicting velocity fields from seismic data.
\par The optimal parameters ($\bm{\theta}$) of the DeepONet are determined by minimizing a loss function ($\mathcal{L}(\bm{\theta})$), which quantifies the difference between the predicted output and the labeled data. The details of this optimization process are discussed further in the numerical section (\Cref{Section:Numerical results: DeepONet}). Mathematically, the optimization problem is expressed as:
\begin{equation}
    \bm{\theta}^* = \min_{\bm{\theta}}\;\mathcal{L}(\bm{\theta}) = \min_{\bm{\theta}}\;\lVert V-\hat{V}(\bm{\theta}) \rVert_{m},
    \label{Eq:Optimization}
\end{equation}
where $V$ represents the true velocity field, $\hat{V}(\bm{\theta})$ denotes the predicted velocity field parameterized by $\bm{\theta}$, and $\lVert \cdot \rVert_m$ is the chosen norm used to measure the discrepancy between the two.
\subsection{Traditional Full waveform inversion with Operator output as initial model}
\label{Subsection:Lagecy FWI}
\begin{figure}[H]
    \centering
    \begin{subfigure}[b]{0.45\textwidth}
        \centering 
        \includegraphics[width=0.95\textwidth, trim={0.5cm, 0.5cm, 0.5cm, 0.5cm}, clip]{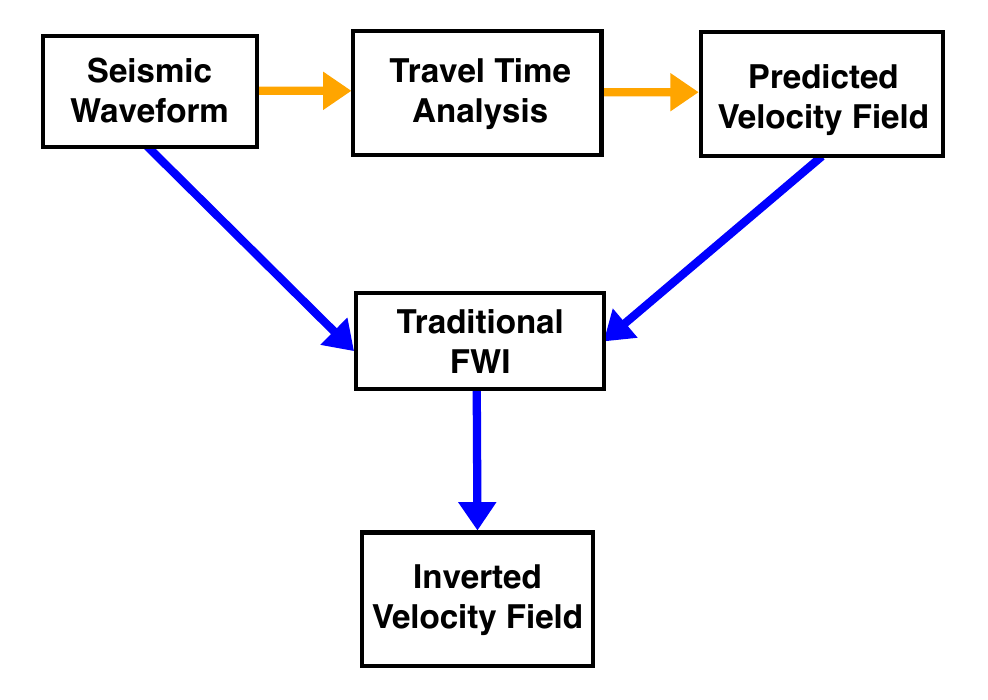}
    \caption{A conventional FWI scheme.}
    \end{subfigure}
    \begin{subfigure}[b]{0.45\textwidth}
        \centering
        \includegraphics[width=0.95\textwidth, trim={0.5cm, 0.5cm, 0.5cm, 0.5cm}, clip]{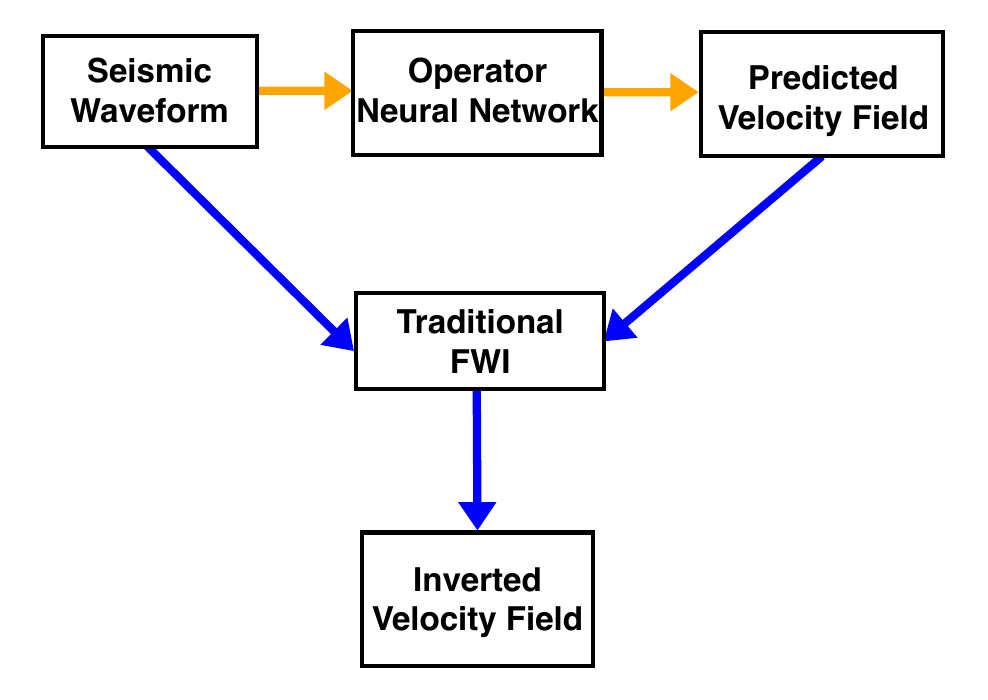}
        \caption{Proposed operator infused hybrid scheme.}
    \end{subfigure}
    \caption{\textbf{Schematic showing a hybrid method of DeepONet and existing FWI.} (a) A schematic showing a conventional FWI scheme, where an initial velocity field is predicted using a method like travel-time analysis for the FWI process. (b) A schematic showing the proposed operator-infused hybrid FWI scheme. In the first step, we predict the velocity field using the recorded seismic waveforms using a trained DeepONet. This predicted velocity field is considered as the informed velocity for existing implementation of FWI workflow, e.g., here we used the method proposed by \citeauthor{virieux2009overview} \cite{virieux2009overview} as inversion traditional FWI}
    \label{Fig:Legacy FWI}
\end{figure}

While neural networks like DeepONet excel at predicting outputs within the training data distribution, their performance tends to degrade significantly for out-of-distribution predictions. Conversely, traditional FWI methods are highly dependent on the initial velocity model, which can lead to suboptimal results if the starting model is inaccurate. To address these limitations, we propose a hybrid approach that leverages DeepONet to generate an initial velocity field, which is then refined using traditional FWI. A schematic diagram of the proposed hybrid method is presented in~\Cref{Fig:Legacy FWI}. For the traditional FWI component, we employ the inversion method described by \citeauthor{virieux2009overview} \cite{virieux2009overview}. 
\par This approach combines the generalization capabilities of DeepONet for generating a plausible initial model with the iterative refinement strengths of traditional FWI. 
\subsection{Training-testing dataset and pre-processing}
\label{Subsection:Data generation}
The dataset used to train DeepONet in this study is sourced from OpenFWI \citep{OpenFWI}, a publicly available benchmark repository designed for large-scale, multi-structural FWI datasets. OpenFWI provides 12 velocity model families encompassing diverse geological structures, such as flat and curved layers, faults, CO$_2$ storage reservoirs, and natural structures, making it a valuable resource for research on FWI. These velocity families are generated using various synthetic velocity models and are tailored for both 2D and 3D FWI studies.
\par In this study, we consider the 2D velocity models from OpenFWI. For each velocity model, the authors solved the forward problem described by \Cref{Eq:Governing equation} for different velocity fields in the same model. The 2D velocity field domain covers an area of \(690 \, \text{m} \times 690 \, \text{m}\) and is discretized into a grid of \(70 \times 70\) points. 
Five seismic sources are placed at evenly spaced intervals along the surface. The resulting seismic waves propagate through the subsurface and are recorded at 70 surface receivers, also evenly distributed along the surface. The recordings are sampled with a time step of $dt = 10^{-3} \, \text{s}$ and extend up to 1000 time steps, yielding a total recording duration of $0.999 \, \text{s}$. For each source, the recorded seismic wavefield has dimensions \(70 \times 1000\), where 70 corresponds to the number of receivers and 1000 represents the number of time measurement. With five sources, the total data size increases to \(70 \times 1000 \times 5\). If \(S\) represents the total number of velocity models for a given velocity family, the complete dataset has a size of \(S \times 70 \times 1000 \times 5\), corresponding to \(S\) velocity fields, where each velocity model is represented by a \(70 \times 70\) grid. 
\par In this study, we use the ``FlatFault-A" velocity family to train our proposed DeepONet architecture. This family consists of 54,000 velocity models representing subsurface structures with 2, 3, and 4 layers, including faults. From this dataset, 48,000 samples are used for training and 6,000 samples are reserved for testing. The training and testing datasets are the same as those utilized by \citeauthor{OpenFWI} \citep{OpenFWI}. 
\par The branch input for DeepONet consists of seismic waveforms. However, the first arrivals in these waveforms typically have much larger amplitudes compared to the later arrivals. This disparity can make it difficult to capture the finer details of the later arrivals. To address this, we apply a logarithmic gain function \citep{OpenFWI}, $\text{log1p}(.)$, which enhances the relative amplitude of the later arrivals. Thus, the pre-processed seismic trace for the DeepONet branch input is given as:
\begin{equation}
    \text{Trace}_{\text{pre-process}} = \text{log1p}(|\text{Trace}|)\times\text{sign}(\text{Trace}),
    \label{Eq:gain function}
\end{equation}
where $\text{log1p}(x) = \log(1 + x)$. We observed that the late arrivals are considerably amplified by the gain function. Furthermore, we normalize the input data between $[-1,1]$ using the maximum and minimum values of the training data set. 
\par \citeauthor{OpenFWI} \cite{OpenFWI} also studied the inverse problem discussed in \Cref{Subsection:Problem statement} using three different neural network architectures for the 2D velocity field: InversionNet \citep{Yue_2020_InversionNet}, VelocityGAN \citep{Zhongping_2020}, and UPFWI \cite{Peng_2021}. In this study, we compare the results of our DeepONet model with those obtained by InversionNet, as reported in \citep{OpenFWI}. InversionNet is a data-driven method that directly approximates the inverse operator for FWI by leveraging a CNN with an encoder-decoder architecture. The encoder extracts high-level features from the input seismic data, while the decoder reconstructs the subsurface velocity model. Additionally, InversionNet incorporates Conditional Random Fields (CRF) to enhance structural details in the velocity models, capturing sharper boundaries and fault features effectively.
\section{Computational experiments: Validation of DeepONet}
\label{Section:Numerical results: DeepONet}
This section focuses on the training process and results of the proposed DeepONet using the training dataset described earlier. The DeepONet was trained using the single-step optimization approach described in \Cref{Eq:Optimization}, and the corresponding numerical results are thoroughly analyzed in this section. We validate the DeepONet architecture under various conditions: with clean noise-free seismic waveforms in \Cref{Subsection:Results DeepONet}, with noisy seismic waveforms in \Cref{Subsection:Noisy traces}, with missing waveform data in \Cref{Subsection:Results DeepONet:missing trace}, and for out-of-distribution predictions in \Cref{Subsection:Result DeepONet:Out of distribution}.
\par The velocity field values in the training dataset range between a minimum of 1500 m/s and a maximum of 4500 m/s. To ensure that the DeepONet predictions remain within these bounds, we apply a scaled sigmoid function to the output. The velocity field is then approximated as
\begin{equation}
\widehat{c}(x,y)\approx \mathcal{G}_{\bm{\theta}}(u)(x,y) = 1490+\text{sigmoid}\left(\mathcal{G}_{\bm{\theta}}(u)(x,y)\right)\times(4510-1490).
\label{Eq:output bound}
\end{equation}
We consider 1490 and 4510 instead of 1500 and 4500, respectively. Since the co-domain of the sigmoid function is  $]0, 1[$. In order to modify the co-domain to $[0, 1]$, thus augmenting the prediction co-domain between $]1490-4510[$. The proposed DeepONet model is trained on the dataset (FlatFault-A) discussed in the previous section. We considered a mean square error (MSE) loss between the labelled data and the predicted velocity model using \Cref{Eq:output bound}. The parameters (weights and biases) of the DeepONet are optimized using the Adam optimizer \citep{Kingma_2014adam} using Tensorflow version 2 in float 32 precision. In order to avoid over-fitting of DeepONet, we consider $L_2$ weight regularizer after some epoch.
\par The accuracy of the prediction for each velocity field (sample) is accessed using two matrices, relative $L_2$ error and structural similarity index measure (SSIM). We consider relative $L_2$ error as the primary error matrix and given for each sample as,
\begin{equation}
    \text{Relative}\;\; L_2\;\;\text{error} = \dfrac{||y - \hat{y}||_2}{||y||_2} = \dfrac{\sqrt{\sum\limits_{j=1}^{n_y} \left(y_{j} -  \hat{y}_{j}\right)^2}}{{\sqrt{\sum\limits_{j=1}^{n_y} y_{j}^2}}}
\end{equation}
where $n_y$ is the number of data points in the velocity field. For global comparison over all the samples, we consider the mean ($\mu$) and standard deviation ($\sigma$) of all the samples along with studying the distribution of the error.
\par We compared our DeepONet results with InversionNet results. The InversionNet results are generated using the trained neural network from \cite{OpenFWI}. \citeauthor{OpenFWI} \cite{OpenFWI}, trained their network using $L_1$ or $L_2$ loss function and observed the error in the case of $L_2$ loss function is smaller. We considered the trained network from \cite{OpenFWI} trained using $L_2$ loss function. 

\subsection{Validation of DeepONet with noise-free data}
\label{Subsection:Results DeepONet}
First, we test our model with noise-free data, i.e., we consider the simulated data as input data without considering any added noise to the recorded seismic waveforms. The mean and standard deviation of relative $L_2$ error in testing are 2.05\% 
\begin{figure}[H]
    \centering
    \includegraphics[width=1.\linewidth, trim={0.325cm 0.325cm 0.325cm 0.2cm},clip]{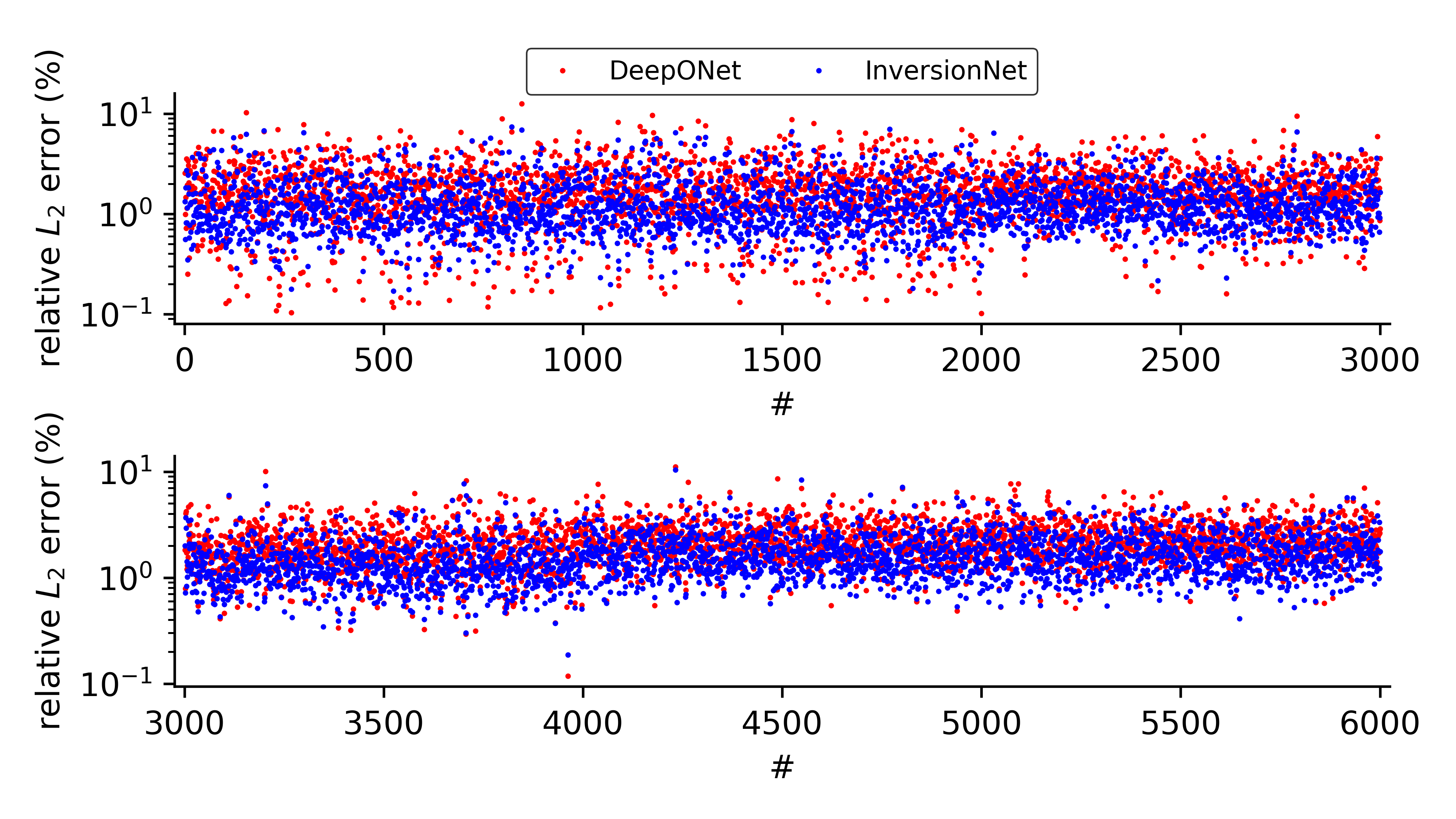}
    \caption{\textbf{Scatter plot of relative $L_2$ error, Clean seismic waveforms:} Scatter plot of relative $L_2$ error for the predicted individual test samples when predicted using DeepONet and InversionNet. The x-axis shows the individual 6000 test samples and the y-axis shows the corresponding relative $L_2$ error. The overall result is comparable. We observed that in many cases, the error for DeepONet is smaller than that of InversionNet and vice versa. The violin plots for the relative $L_2$ error are shown in \Cref{Fig:Clean data:Violin}, showing the qualitative distribution of the relative $L_2$ error.}
    \label{Fig:Clean data:Scatter plot}
\end{figure}
\begin{wrapfigure}{r}{9cm}
\centering
\includegraphics[width=8.5cm]{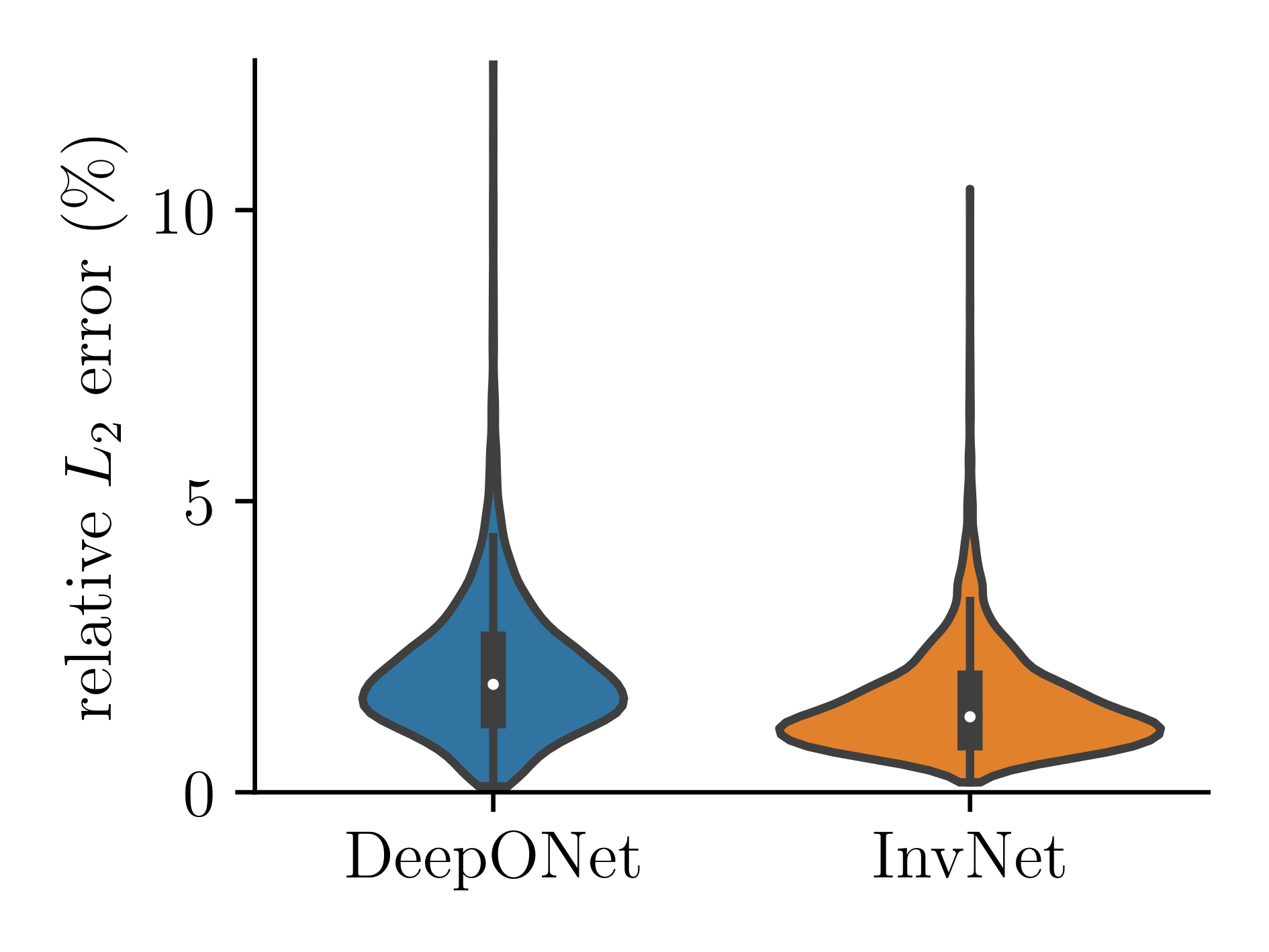}
\caption{\textbf{Violin plot of relative $L_2$ error, Clean seismic waveforms:} Violin plot of relative $L_2$ error for predicted velocity model for test samples when predicted using DeepONet and InversionNet (InvNet).} 
\label{Fig:Clean data:Violin}
\end{wrapfigure}
\noindent and 1.14\%, respectively. Compared with InversionNet results from \cite{OpenFWI} (with mean 1.53\% and standard deviation 0.89\%), the DeepONet results show a slightly higher mean error and larger standard deviation. In \Cref{Fig:Clean data:Scatter plot}, we have shown the scatter plot of relative $L_2$ error for the test samples when predicted using DeepONet and InversionNet. We observed that in many cases, the error in DeepONet is smaller than InversionNet and vice-versa. Next, we show the violin plots of distribution of relative $L_2$ error in \Cref{Fig:Clean data:Violin}, computed for the test samples corresponding to DeepONet and InversionNet architectures. It is to be noted that the violin plot of DeepONet is slightly wider near zero than InversionNet, indicating more samples near zero. Furthermore, the violin plot for DeepONet has a long tail toward higher error, indicating skewness of $L_2$ error for DeepONet, which along with more sample near zero, causes the higher  standard deviation for DeepONet compared to InversionNet.
\par
\begin{figure}[H]
    \centering
    \includegraphics[width=1.0\textwidth, trim={0.15cm 0 0.15cm -0.25cm},clip ]{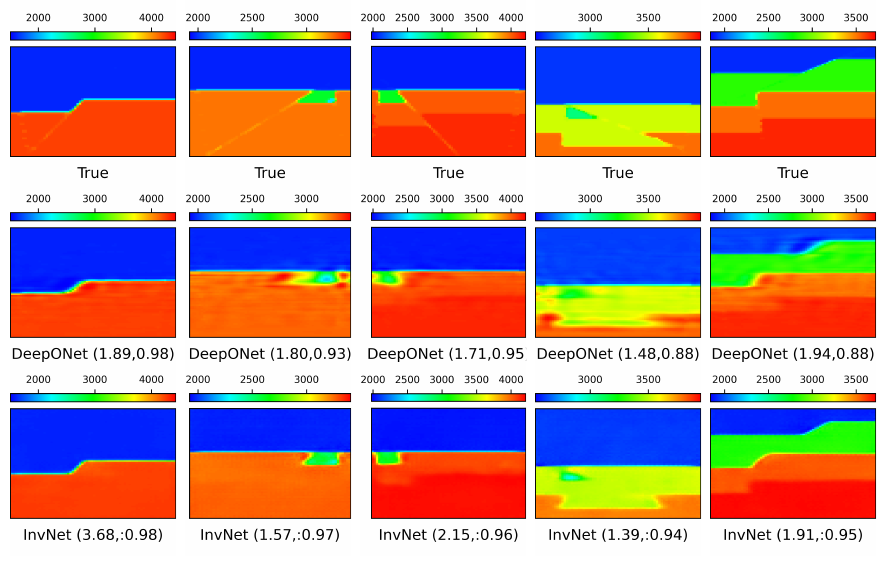}
    \caption{\textbf{A few predicted velocity fields, Clean seismic waveforms:} A few predicted velocity fields when predicted using DeepONet on the test dataset compared against predictions from InversionNet. Each column in the first row shows the true velocity field, and the second and third rows show the predicted velocity fields obtained from DeepONet and InversionNet (InvNet), respectively. In the figure, the two numbers indicated in the bracket (x.xx, x.xx) in the second and third rows show the \% relative $L_2$ error and SSIM of the predicted velocity field with respect to the true velocity field.}
    \label{Fig:Clean data:DeepONet vs InvNet}
\end{figure}

 In \Cref{Fig:Clean data:DeepONet vs InvNet}, we have shown a few representative samples of the predicted velocity field using both DeepONet and InversionNet. The two numbers in the bracket below the predicted velocity field show the \% relative $L_2$ error and SSIM between the true and predicted velocity fields. We observed that both methods show relatively good accuracy with the true velocity field. The distinctions between the layers are clearly visible in the predicted velocity field, similar to the true velocity field, except for the third representative sample. In the case of the third sample, the contrast between the second and third layers is small, which may be the reason for the inability to predict the layers clearly. We also observed that the InversionNet is better predicting the fault plane as observed in second and third samples.
\subsection{Validation of DeepONet with noisy data}
\label{Subsection:Noisy traces}
In the previous section, we discussed the results of DeepONet for noise-free data. However, in practical scenarios, the data are often noisy for various reasons, e.g. interference from other seismic sources . Thus, we tested our noise-free trained DeepONet model using noisy data and compared these results with InversionNet's (noise-free trained) performance. 
\par To add noise, we added filtered Gaussian noise to the simulated seismic waveforms of FlatFault-A for the test velocity fields. We added a filtered Gaussian noise for each simulated trace as follows. For a seismic waveform $X_{ij}$ (shape $(n_t\times 1)$) for a particular source $i$ and station $j$, we generated a Gaussian noise of the same size of ($n_t\times1$) as that of $X_{ij}$ with a standard deviation $\sigma$. Then, the generated noise is filtered to remove the high frequencies, with a cut of frequency of 100. The filtered noise is amplified by multiplying it with the maximum value of the seismic waveform ($\max(X_{ij})$). This amplified noise is added to the original seismic waveform $X_{ij}$. We add noise to all the simulated traces for all the stations and sources. Furthermore, as discussed earlier, we considered the logarithmic gain function function and normalization of the seismic waveforms.
\par
\begin{figure}[H]
    \centering
    \includegraphics[width=1\textwidth]{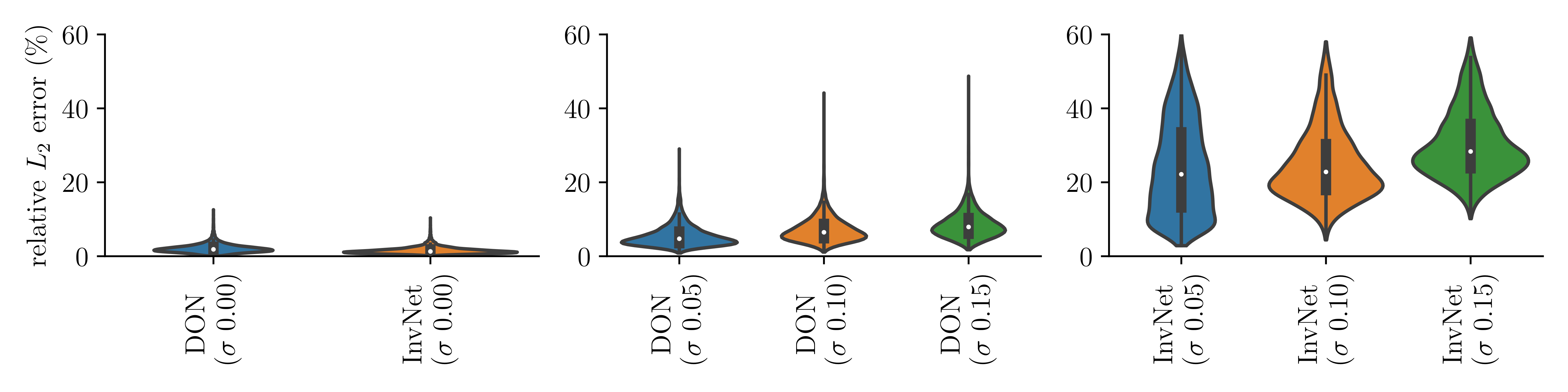}
    \caption{\textbf{Violin plot of relative $L_2$ error, noisy data:} Violin plot of relative $L_2$ error for predicted velocity field for test samples when predicted using DeepONet (DON) and InversionNet (InvNet) for different standard deviation of the added Gaussian noise. The plots are plotted in three subplots, keeping the $y$ axis the same. This is done to better visualize the plots; otherwise, violin plots for the InversionNet with noise will look much slandered compared with the noise-free (clean) and DeepONet plots. The first plot shows the violin plots for DeepONet and InversionNet when predictions are made with clean seismic waveforms. The second plot shows the violin plots for DeepONet when predictions are made with noisy seismic waveforms of standard deviation $\{0.05, 0.10, 0.15\}$. The third plot shows the violin plots for InversionNet when predictions are made with noisy seismic waveforms of standard deviation $\{0.05, 0.10, 0.15\}$. It can be observed (along with \Cref{Table:Noisy data: Mean and SD}) that DeepONet results are better than InversionNet when predictions are made with noisy seismic waveforms.}
    \label{Fig:Noisy data:Violin plot}
\end{figure}

 We predict the velocity field for noisy data for different standard deviations of the noise. For each standard deviation, we predicted 6000 test velocity fields. The mean and standard deviation of the relative $L_2$ error for the test samples are shown in  \Cref{Table:Noisy data: Mean and SD}, and the violin plots of the relative $L_2$ error are shown in \Cref{Fig:Noisy data:Violin plot}. The mean and standard deviation of relative $L_2$ error is much smaller in the case of DeepONet prediction than InversionNet. In the case of DeepONet, the mean error increased from 2.05\% for noise-free data to 5.44\% when noise with a standard deviation of 0.05 was added to the seismic waveforms. However, for InversionNet, the mean value increased from 1.53\% for clean seismic waveforms to 23.96\% for the same case. At higher levels of noise, errors for both increased, with the DeepONet errors being consistently lower than InversionNet's errors, demonstrating that the proposed DeepONet architecture is less sensitive to noise than InversionNet.
\begin{table}[H]
\centering
\caption{\textbf{Prediction with noisy data:} Mean and standard deviation ($\mu, \sigma$) of relative $L_2$ error for the predicted velocity field when predicted using noisy seismic waveforms evaluated using DeepONet and InversionNet under different values of the standard deviation of the added Gaussian noise.}
\label{Table:Noisy data: Mean and SD}
\begin{tabular}{L{4.75cm}|C{2cm}C{2cm}C{2cm}C{2cm}}
\hline
Standard deviation of added noise  & $0$ & $0.05$ & $0.10$ & $0.15$ \\ \hline
DeepONet (\%) & (2.05, 1.14) & (5.44, 2.73) & (7.10, 3.13) & (8.47, 3.44)\\ \hline
InversionNet (\%) & (1.53, 0.89) & (23.96, 12.70)  & (24.91, 9.66) & (30.30, 9.05)\\  \hline
\end{tabular}
\end{table}
\par
\begin{figure}[H]
    \centering
    \begin{subfigure}[b]{1\textwidth}
    \centering
    \includegraphics[width=1\linewidth]{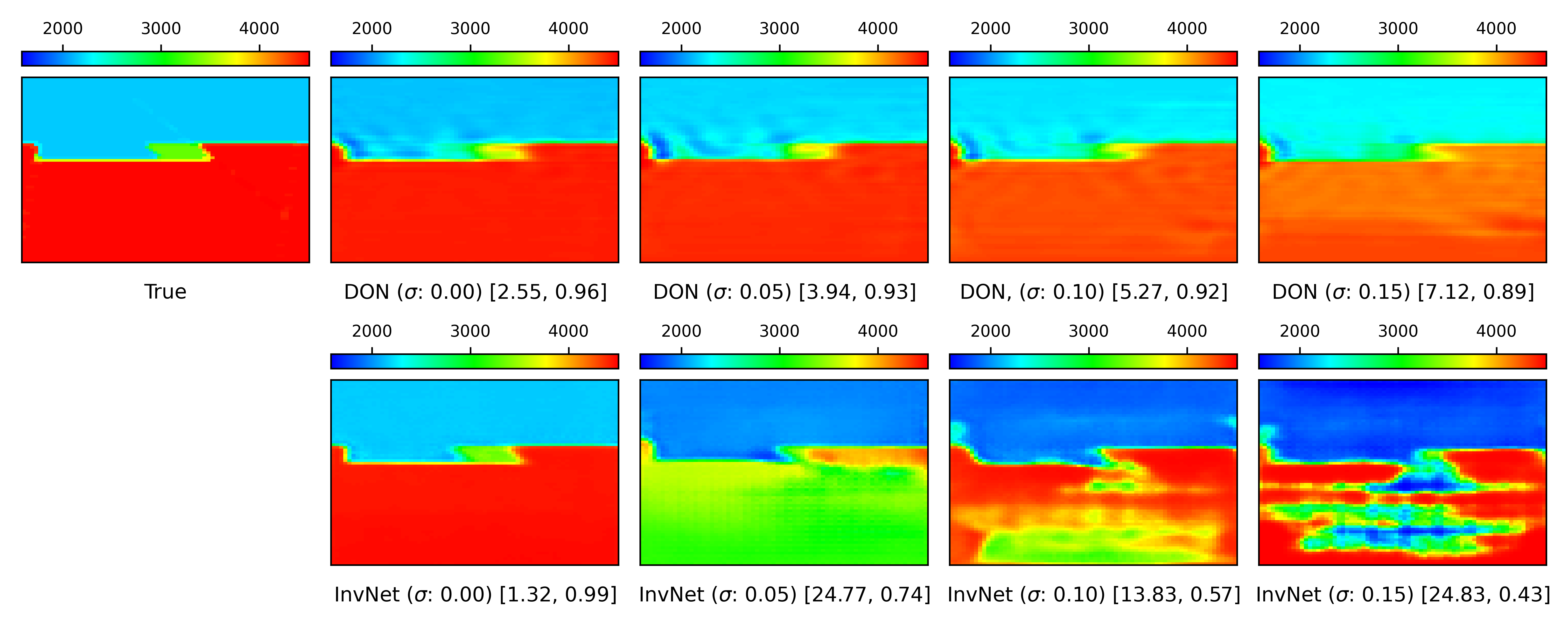}  
    \caption{\textbf{Sample velocity field 1}}
    \end{subfigure}
    \vskip\baselineskip
    \begin{subfigure}[b]{1\textwidth}
    \centering
    \includegraphics[width=1\linewidth]{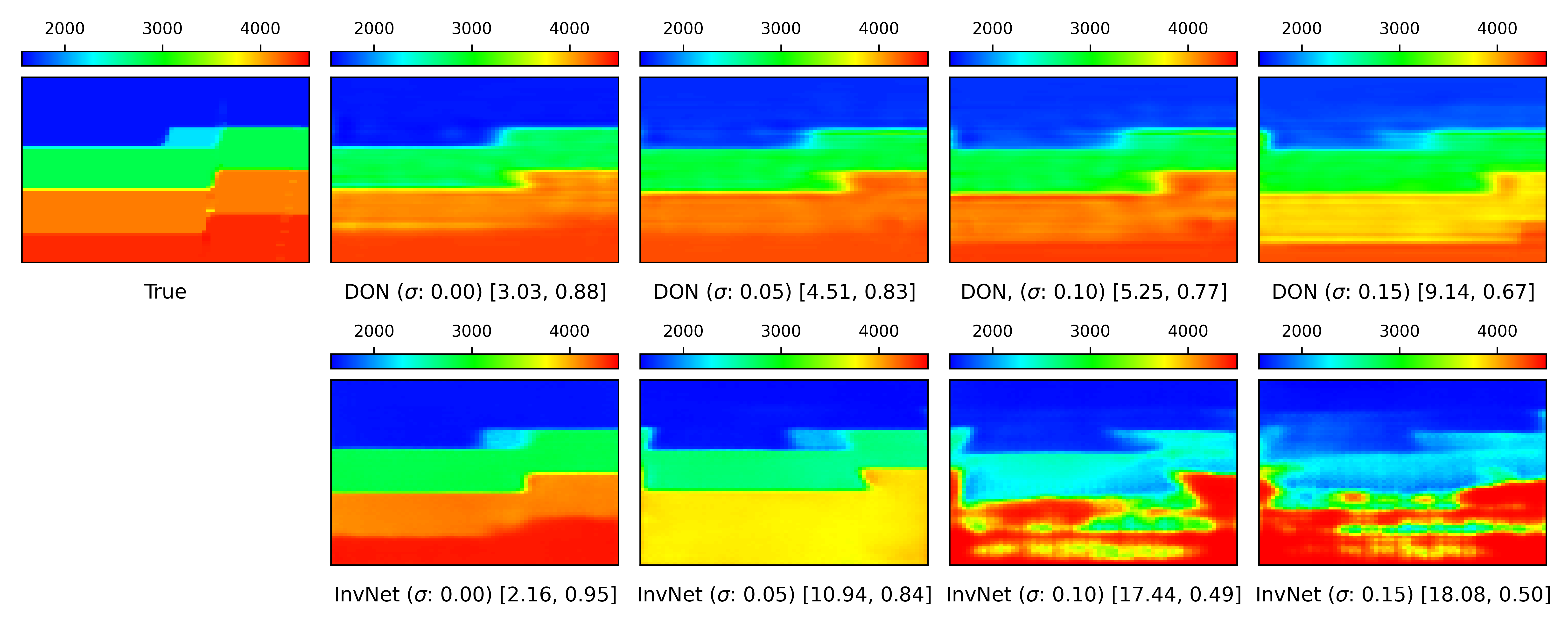}
    \caption{\textbf{Sample velocity field 2}}
    \end{subfigure}
    \caption{\textbf{Predicted velocity field, noisy data:} Two predicted velocity fields when predicted using DeepONet (DON) and InversionNet (InvNet) on the test dataset. The predictions are carried out with noisy seismic waveforms. The first column of the first row shows the true velocity field. The second column of the first and second rows shows the predicted velocity fields when predicted using DeepONet and InversionNet with clean seismic waveforms. The third to fifth columns of the first row shows DeepONet's predicted velocity field with noisy seismic waveforms with standard deviations of $\{0.05, 0.10, 0.15\}$ of the added Gaussian noise respectively. The corresponding predictions using InversionNet are shown in the third to fifth columns of the second row, respectively.}
    \label{Fig:Noisy data:Prediction}
\end{figure}

 We have shown two representatives predicted test samples of velocity fields in \Cref{Fig:Noisy data:Prediction} for different values of the standard deviation of the added Gaussian noise. The \% relative $L_2$ error in InversionNet prediction is higher than DeepONet prediction when Gaussian noise is added. It can be observed that DeepONet preserves most of the structure of the true velocity field even for the higher noise levels ($\sigma=0.15$), while InversionNet result introduces artificial structure. In the case of a low noise ($\sigma=0.05$), InversionNet preserved the structure of the true velocity field; however, the velocity offsets from its true value, as seen in the case of sample velocity field 1. Thus, we would like to conclude that though the mean and standard deviation of relative $L_2$ error is higher in the case of DeepONet than InversionNet when predicted using clean seismic waveforms, the mean and standard deviation of relative $L_2$ error is smaller in the case of DeepONet than InversionNet when predicted using noisy seismic waveforms. Therefore, DeepONet is a more practical choice than InversionNet for predicting velocity fields for field studies which inevitably have noise. A rigorous mathematical or theoretical analysis for why InversionNet has higher errors than DeepONet for noisy data is beyond the scope of the present study. However, a heuristic argument can be made as follows: In the case of InversionNet, the encoder and decoder are part of a single network where the encoder network is followed by the decoder network. The output of the encoder network is the input to the decoder network. Thus, if a noisy input seismic waveform is considered, it affects the encoder part of the network, and the effect propagates to the decoder network. However, in the case of DeepONet, the encoder (the branch) and decoder (the trunk) are two separate networks. While noisy input seismic waveforms affect the branch networks (encoder), the decoder network (trunk) is not affected by the noisy input seismic waveforms. In other words, in the case of DeepONet, the basis functions given by the trunk network are not affected by noise; only the coefficients are affected by the noise.
\subsection{Validation of DeepONet with missing stations}
\label{Subsection:Results DeepONet:missing trace}
In the previous sections, we discussed the numerical results of DeepONet with clean and noisy seismic waveforms and compared them with InversionNet. Apart from noise, another practical difficulty in data acquisition in the field is the inability to place stations in certain locations. Thus, we investigate the performance of DeepONet and InversionNet for the case of missing receivers. For this purpose, we kill the seismic traces corresponding to five receivers for all the sources. We also consider a case combining missing and noisy data. We have observed in the previous section that InversionNet is more sensitive to noisy seismic waveforms, and error is higher even for a standard deviation of 5\%. Thus, in the case of missing seismic waveforms with noise, we have considered the standard deviation of the added Gaussian noise to be only 2\%. Furthermore, as discussed earlier, we have considered the log gain function and normalisation for the seismic waveforms. 
\par
\begin{figure}[H]
    \centering
     \includegraphics[width=1\textwidth]{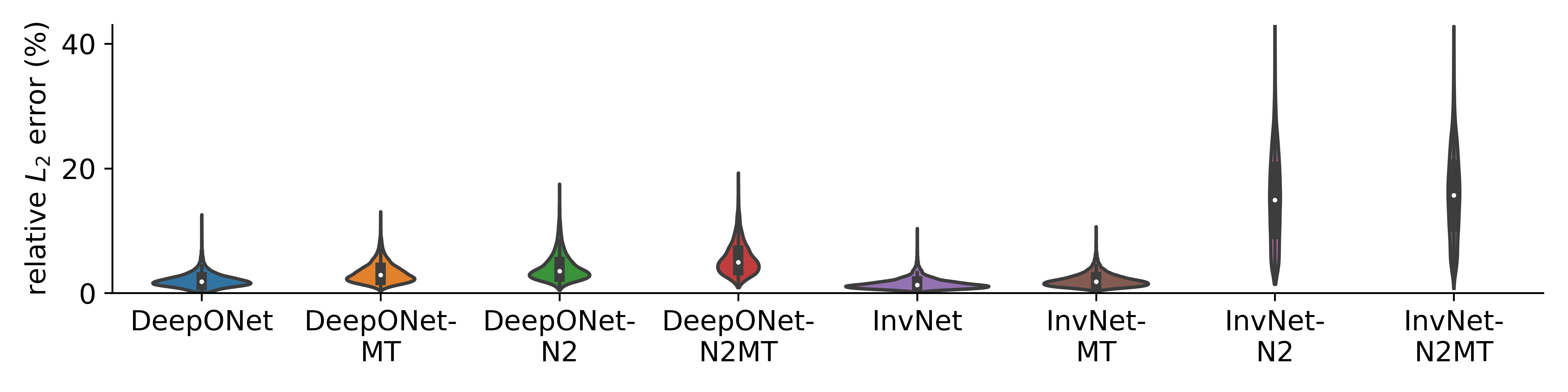}
    \caption{\textbf{Violin plot of relative $L_2$ error, missing seismic waveforms:} Violin plot of relative $L_2$ error for the predicted velocity field for test samples when missing seismic waveforms is considered. Different cases are considered as \Cref{Table:Missing trace results}.}
    \label{Fig:Violin plot missing trace}
\end{figure}

 We predicted velocity fields with different conditions as discussed above. We predicted 6000 test velocity fields for each case. The mean and standard deviation of the relative $L_2$ error for the test samples are shown in \Cref{Table:Missing trace results}. The corresponding violin plots for relative $L_2$ error are shown in \Cref{Fig:Violin plot missing trace}. We observed that for both the methods, DeepONet and InversionNet, the error marginally increased when the missing receiver is considered. Furthermore, the effect of noise in the seismic waveform on the accuracy of the predicted velocity field is greater than that of the missing receiver case.
\begin{table}[H]
\centering
\caption{\textbf{Prediction with missing receivers:} The mean and standard deviation ($\mu$, $\sigma$) of relative $L_2$ error in test prediction for different cases of missing seismic waveforms when predicted using DeepONet and InversionNet.}
\label{Table:Missing trace results}
\begin{tabular}{L{3cm}L{8cm}C{2.75cm}}
\hline
\multicolumn{1}{c}{Case} & \multicolumn{1}{c}{Description} & \multicolumn{1}{c}{Relative $L_2$ error (\%) } \\ \hline
DeepONet & DeepONet with clean data and no missing data & (2.05, 1,14) \\ 
DeepONet-MT & DeepONet with clean data and missing data & (3.24, 1.60) \\ 
DeepONet-N2 & DeepONet with noisy data (2\%) and no missing data & (4.08, 2.11) \\ 
DeepONet-N2MT & DeepONet with noisy data (2\%) and missing data & (5.47, 2.51) \\ \hline
InvNet & InversionNet with clean data and no missing data & (1.53, 0.89) \\ 
InvNet-MT & InversionNet with clean data and missing data & (2.07, 1.06) \\ 
InvNet-N2 & InversionNet with noisy data (2\%) and no missing data & (15.1, 7.22) \\ 
InvNet-N2MT & InversionNet with noisy data (2\%) and missing data & (15.8, 6.91) \\ \hline
\end{tabular}
\end{table}
\par In \Cref{Fig:Prediction:Missing trace}, we have shown four representative predicted test velocity fields when predicted with missing data. It can be observed that error marginally increases when missing receivers are considered for both DeepONet and InversionNet. Furthermore, the noisy data has more effect on the error than the missing data. We consider missing data corresponding to 5 sensors near the first half of the length. Future studies may include studies considering the effects of missing traces from different locations.
\begin{figure}[H]
    \centering
    \begin{subfigure}{1\textwidth}
    \centering
     \includegraphics[width=0.95\textwidth]{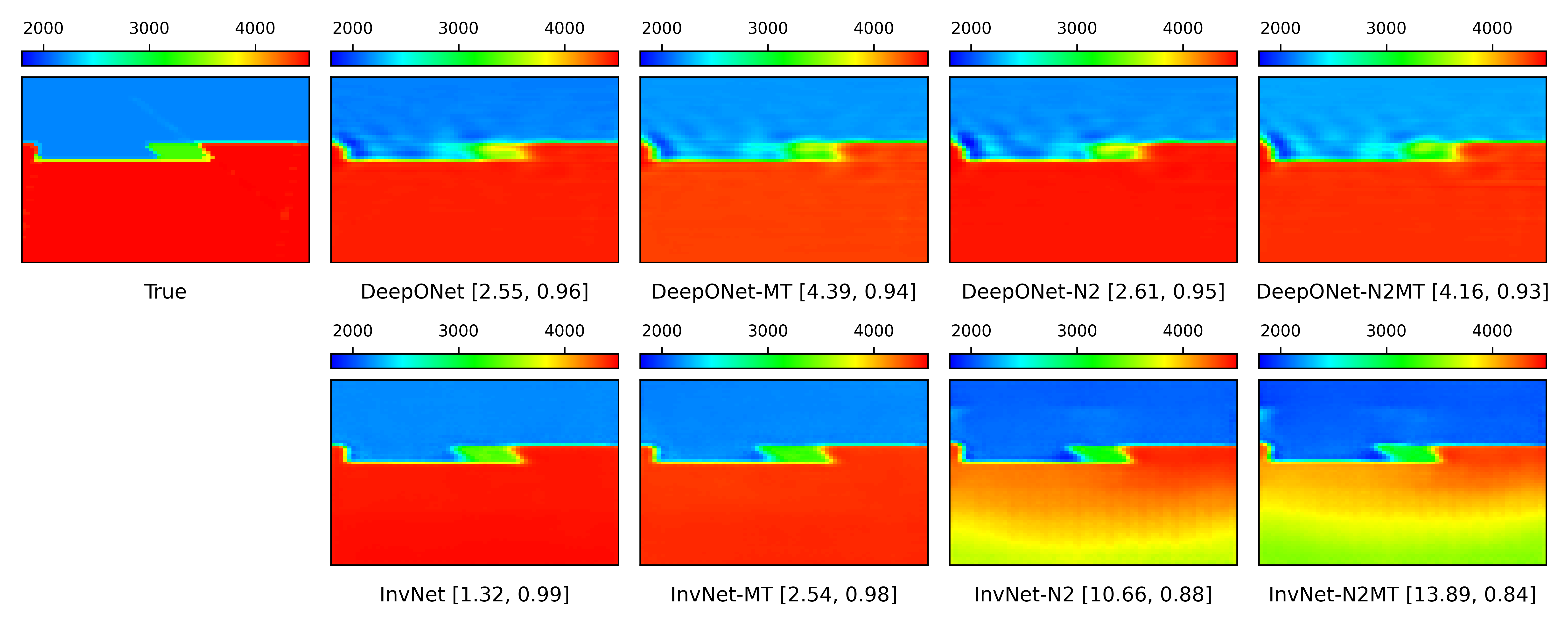}
     \caption{\textbf{Sample velocity field 1}}
    \end{subfigure}
\end{figure}
\begin{figure}[H]\ContinuedFloat
\centering
     \begin{subfigure}{1\textwidth}
    \centering
     \includegraphics[width=0.95\textwidth]{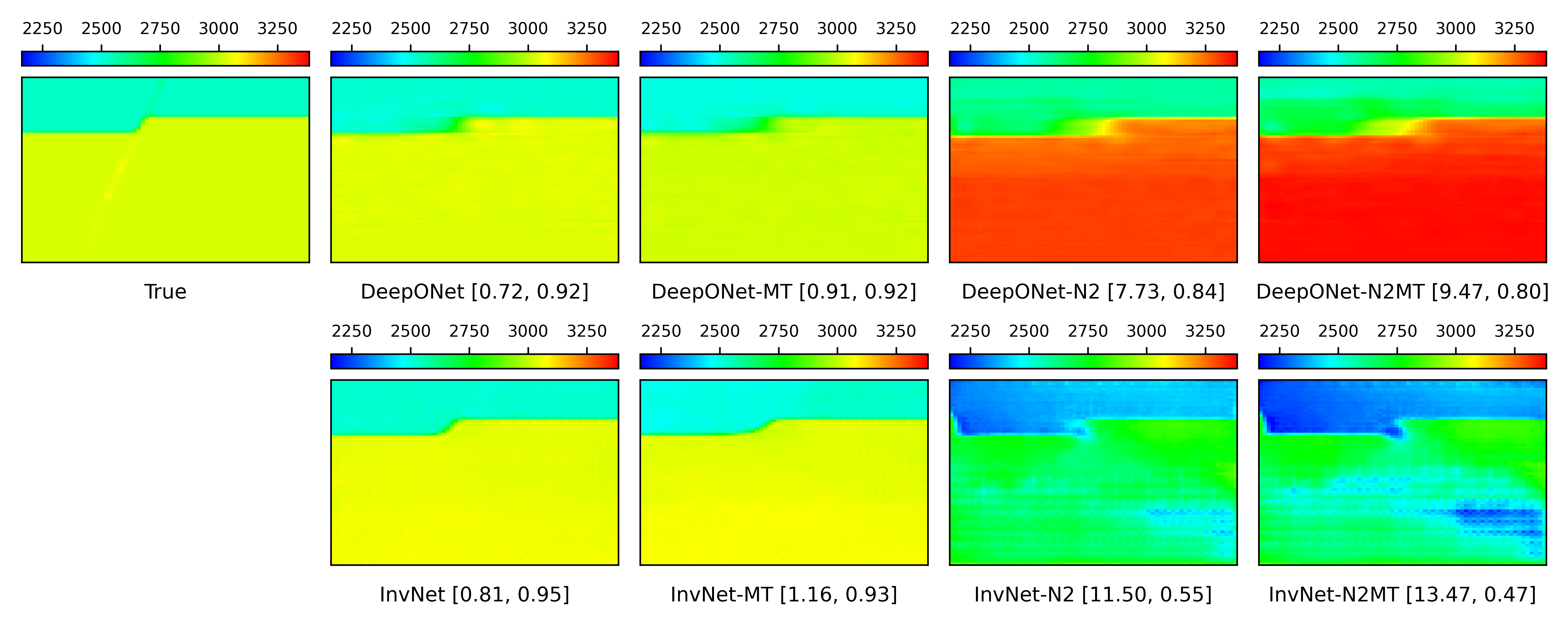}
     \caption{\textbf{Sample velocity field 2}}
    \end{subfigure}
\end{figure}
\begin{figure}[H]\ContinuedFloat
     \begin{subfigure}{1\textwidth}
    \centering
     \includegraphics[width=0.95\textwidth]{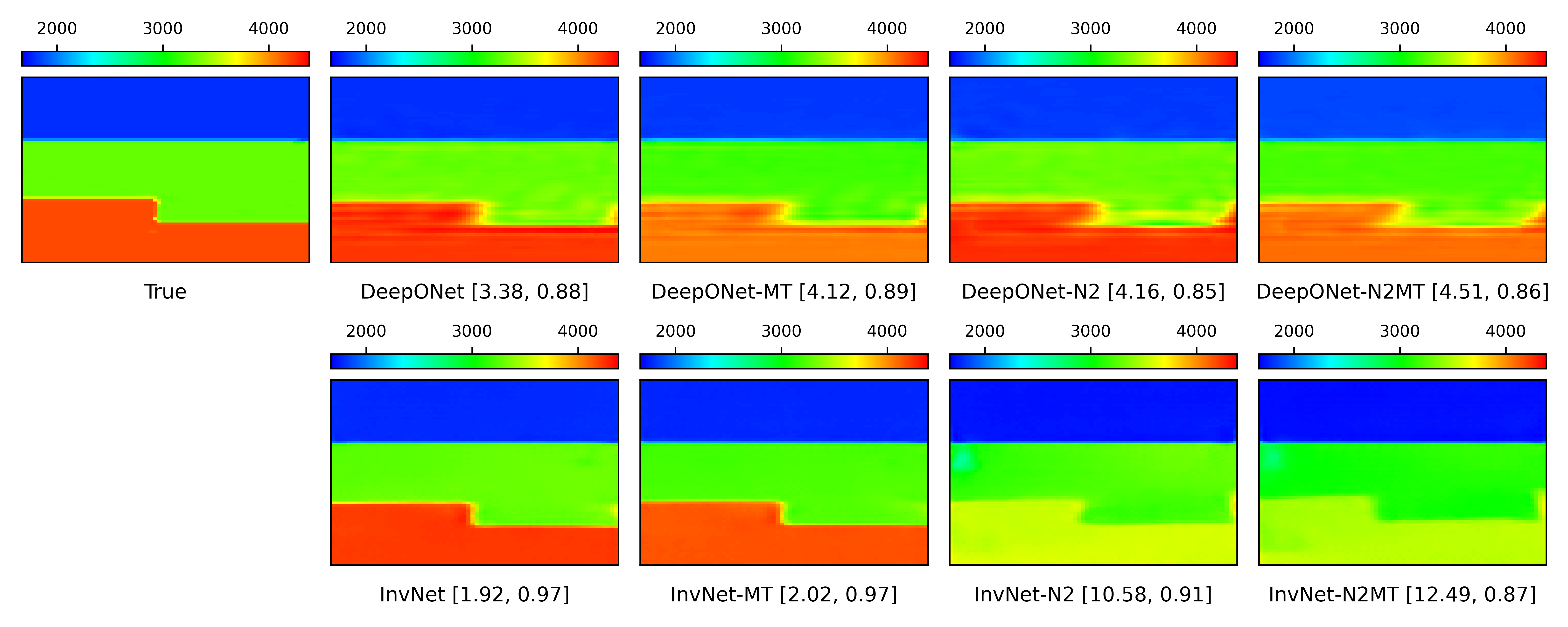}
     \caption{\textbf{Sample velocity field 3}}
    \end{subfigure}
\end{figure}
\begin{figure}[H]\ContinuedFloat
    \centering
     \begin{subfigure}{1\textwidth}
    \centering
     \includegraphics[width=0.95\textwidth]{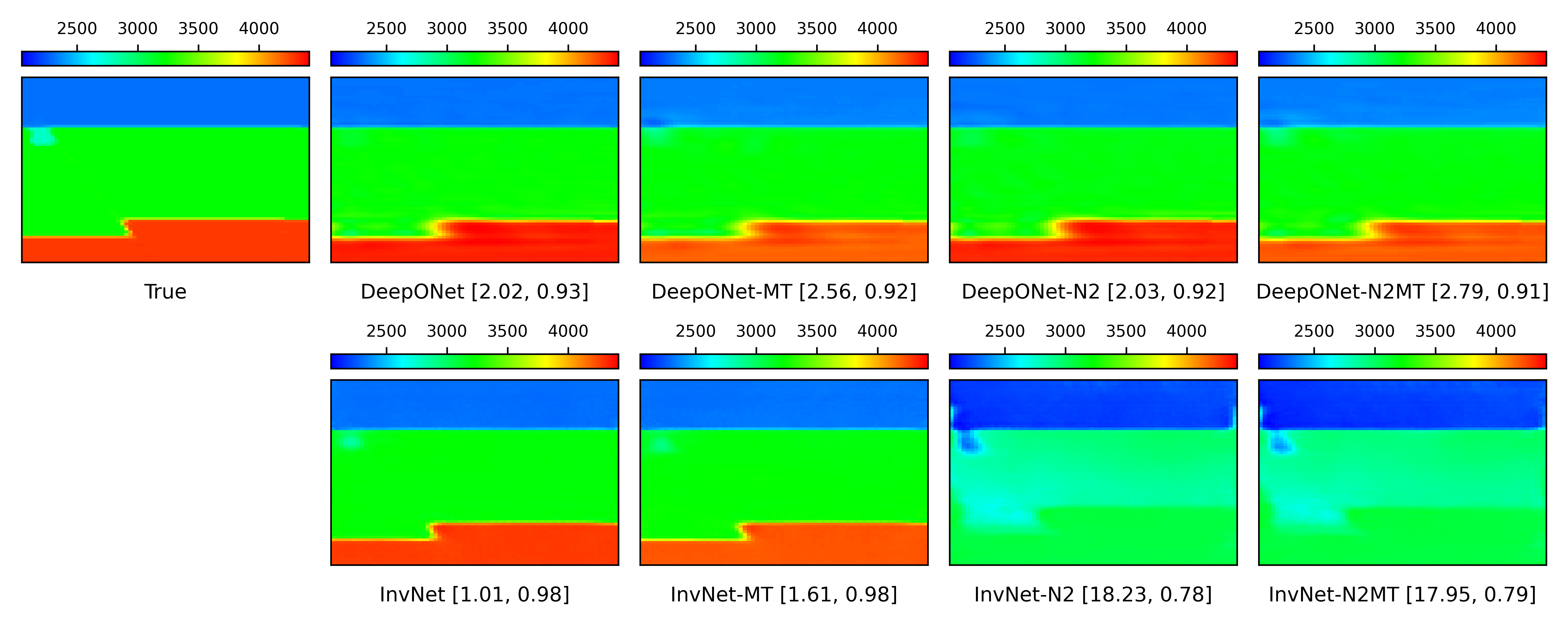}
     \caption{\textbf{Sample velocity field 4}}
    \end{subfigure}
    \caption{\textbf{Prediction with missing receivers:} Four representative samples of predicted test velocity field when predicted using missing receivers data. The first column shows the true velocity field. From second to fifth column show the predicted velocity field, the first row for DeepONet and second row for InversionNet. The four different cases considered are as shown in \Cref{Table:Missing trace results}, i.e. noise free, no missing data (second column), missing receiver data (third column), noisy data (2\%), no missing data (fourth column), noisy data (2\%) and missing receiver data (fifth column). The error in prediction marginally increases when missing receiver is considered.}
    \label{Fig:Prediction:Missing trace}
\end{figure}
\subsection{Validation of DeepONet: Forward Modeling}
\label{Subsection:Forward Modeling}
To validate the velocity model inferred from DeepONet, we conducted forward seismic modeling experiments and compared the particle velocity time histories derived from both the inferred and true velocity models. We solved the system of first-order hyperbolic PDE governing acoustic wave propagation in the medium using the discontinuous Galerkin finite element method. Specifically, we use the method proposed by \cite{chan2018weight, shukla2020weight}. We solved the following wave equation, 
\begin{align}\label{acous-sys}
\begin{aligned}
\frac{1}{\rho c^2(x, y)} \frac{\partial p}{\partial t}=\nabla \cdot \boldsymbol{u} , \\
\rho \frac{\partial \bm{u}}{\partial t}=\nabla p + \bm{f},
\end{aligned}
\end{align}
where $\rho$ is the density of the medium and is taken as 1 for numerical simulation. $c(x, y)$ is the acoustic velocity of the medium. $p$ and $\bm{u}$ are instantaneous pressure and particle velocity vectors, respectively. To perform the numerical simulation of \Cref{acous-sys}, we added a compactly supported forcing function $\bm{f}(\bm{x},t)=[0,~\delta(\bm{x}-\bm{x}_0)h(t)]^\top$, with $h(t)$ as Ricker wavelet
\begin{equation}
h(\boldsymbol{x}, t)=\left(1-2\left(\pi f_0\left(t-t_0\right)\right)^2\right) e^{-\left(\pi f_0\left(t-t_0\right)\right)^2}\delta(\bm{x} -\bm{x}_0),
\end{equation}
where $\delta(\bm{x} -\bm{x}_0)$ is the delta function and $f_0$ is central frequency and considered as 25 Hz with $t_0 = 1/f_0$.

\par
\begin{figure}[H]
    \centering
     \includegraphics[width=1\textwidth]{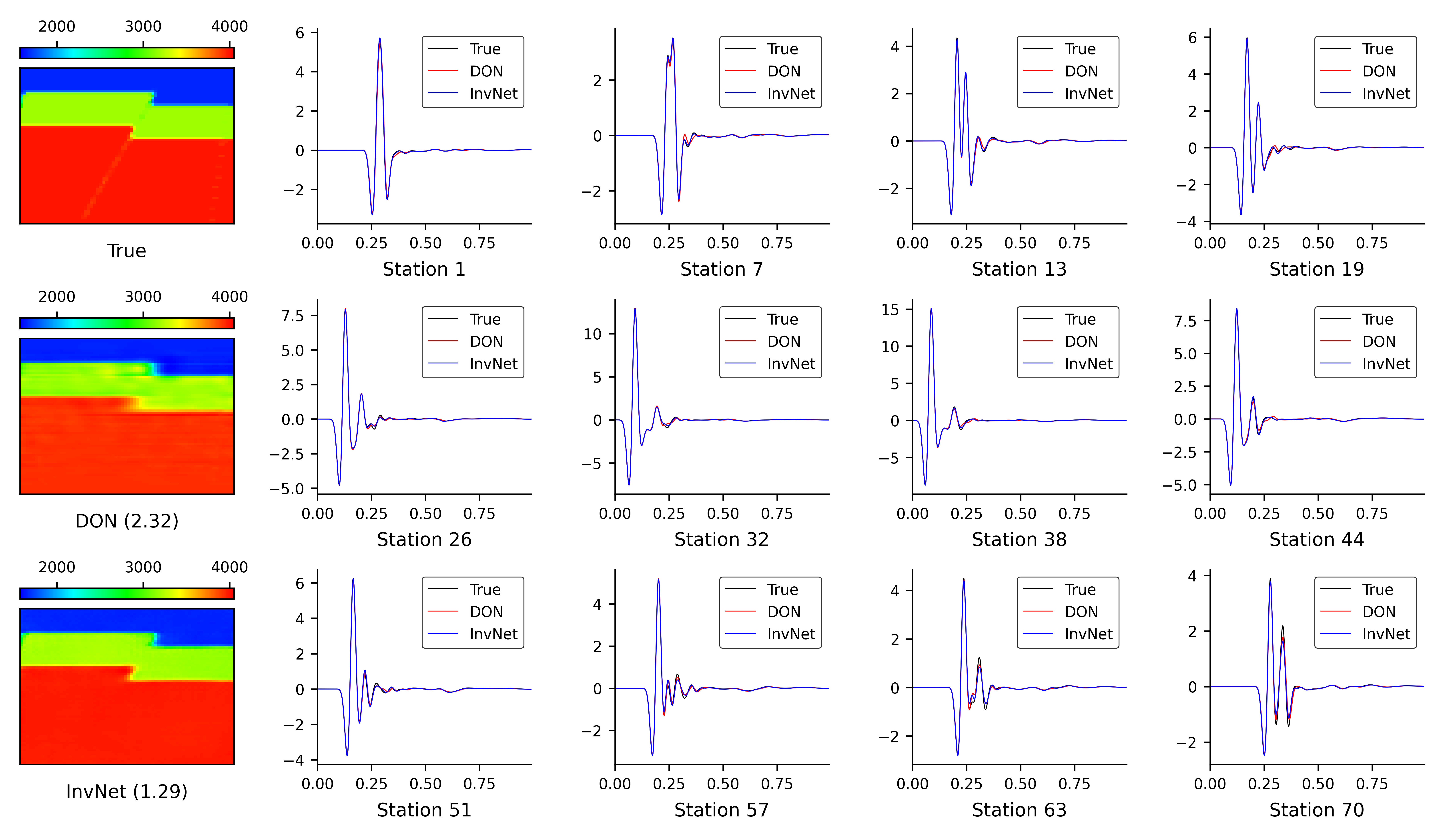}
    \caption{\textbf{Forward model:} Calculated seismic waveforms at different stations. Seventy stations are placed at the surface from left to right, and we have shown traces recorded at 12 stations. In the first column, we have shown the true and predicted velocity model using DeepONet (DON) and InversionNet (InvNet) from top to bottom, respectively. In columns two to five and the first, second and third rows, we have shown the seismic waveforms calculated for the true, DeepONet and InversionNet velocity fields. The plots demonstrate a strong agreement for events near the first break (e.g., Station 1: $t=[0.22~\text{s} - 0.375 ~\text{s}]$) and reasonable alignment in the coda portion of the trace (e.g., Station 1: $t=[0.375~\text{s} - 0.75~\text{s}]$), although the overall global error in the inferred velocity from DeepONet is higher compared to InversionNet}
    \label{fig:acous_model}
\end{figure}

 In \Cref{fig:acous_model}, we show the comparison of traces computed for inferred and true velocity models. The first column in  \Cref{fig:acous_model} shows the true velocity, followed by the velocity inferred from DeepOnet and InversionNet, from top to bottom, respectively. The second, third, fourth and fifth columns represent the plots showing comparisons between time histories of vertical particle velocity at different surface stations. The plots demonstrate a strong agreement for events near the first break and reasonable alignment in the coda portion of the trace, although the overall global error in the inferred velocity from DeepONet is higher compared to InversionNet.

\subsection{Validation of DeepONet: Out-of distribution prediction}
\label{Subsection:Result DeepONet:Out of distribution}
\begin{figure}[H]
    \centering
    \includegraphics[width=1\textwidth]{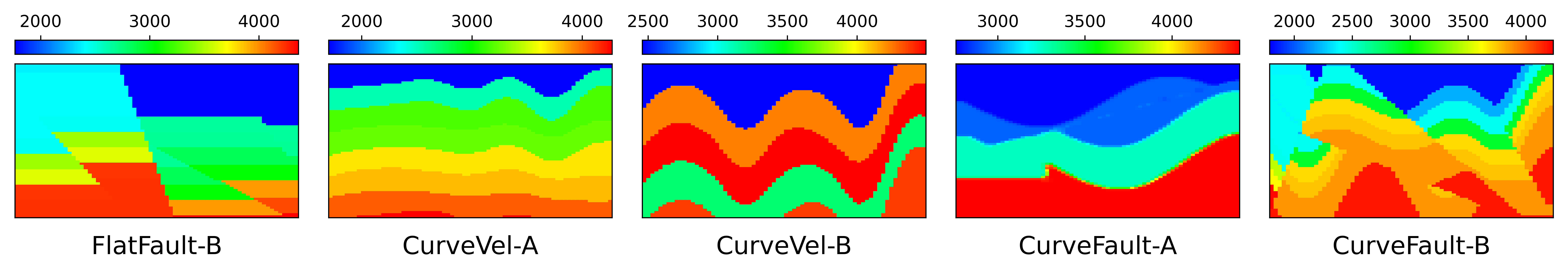}
    \caption{Representative sample for different subsurface velocity models considered for out-of-distribution prediction.}
    \label{Fig:Appendxi:Out of dist vel}
\end{figure}

In this section, we discuss the prediction of velocity fields from other velocity models when trained with the "FlatFault-A" model dataset. The FlatFault-A family of models is a family of subsurface velocity model with flat faults consisting of two, three or four-layered subsurface. After training our DeepONet model on Flatfault-A, we compared predictions for other velocity models, i.e. FlatFault-B, CurveVel-A, CurveVel-B, CurveFault-A, and CurveFault-B. FlatFault-B is the velocity model similar to FlatFault-A but with a more complex and larger number of subsurface layers. The CurveVel-A and CurveVel-B velocity models are without any fault, but the interfaces in  models are curved. The latter one is more complex than the former one. The CurveFault-A and CurveFault-B velocity models are with faults and curve velocity profiles. The latter one is more complex than the previous one. We have shown a representative sample from each velocity model in \Cref{Fig:Appendxi:Out of dist vel}. The domain size of these velocity models is the same as that of the FlatFault-A model ($690\;\text{m}\times 690\;\text{m}$) and has the same mesh size of $70\times 70$. The seismic sources and sensors are the same and placed at the same locations as FlatFault-A. The time steps and total duration of the seismic waveforms are consistent, ensuring that the branch input shares the same characteristics and domain. If the branch input differs in characteristics, predicting the velocity field may become infeasible, or it may be necessary to project the input to match the required characteristics. Similar to previous section, the results obtained for DeepONet are compared to  the InversionNet model \citep{OpenFWI} trained on FlatFault-A, and also tested on the other velocity models.
\par
\begin{figure}[H]
    \centering
    \includegraphics[width=1\linewidth]{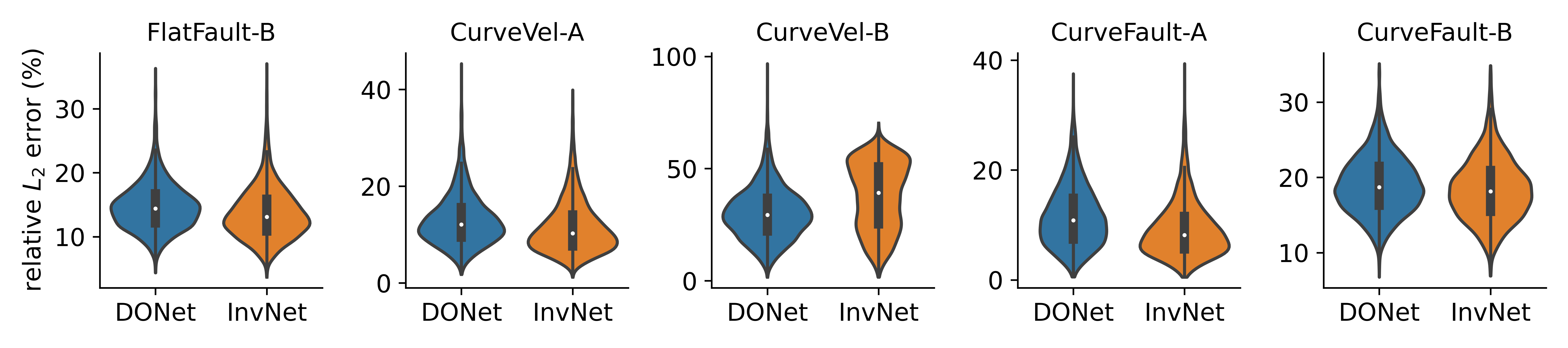}
    \caption{\textbf{Violin plot of relative $L_2$ error, out of distribution:} Violin plot of relative $L_2$ error for the predicted out of distribution velocity field for test samples when predicted using DeepONet (DONet) and InversionNet (InvNet). The plots show similar accuracy for both the DeepONet and InversionNet except CurveVel-B. In the case of CurveVel-B, the plot due to DeepONet is somewhat symmetric with a tail toward higher error. On the other hand, the plot due to InversionNet is skewed towards zero. Furthermore, the mean and standard deviation in the case of InversionNet are higher in the case of InversionNet as observed in \Cref{Table:Out of distribution}}
    \label{Fig:Out of dist:Violin plot}
\end{figure}

 We predict 6000 velocity fields from each velocity model using the trained DeepONet using FlatFault-A. We also predict the velocity fields using InversionNet trained on FlatFault-A (from \cite{OpenFWI}). The mean and standard deviation of the relative $L_2$ errors for different velocity models are shown in \Cref{Table:Out of distribution}, and the corresponding violin plots are shown in \Cref{Fig:Out of dist:Violin plot}. The mean and standard deviation for models FlatFault-B, CurveVel-A, and CurveFault-B are comparable. However, the mean and standard deviation for CurveVel-B show the DeepONet prediction is much more accurate than InversionNet. In the case of CurveFault-A, the mean and standard deviation of the error is smaller in the case of InversionNet than DeepONet.
\begin{table}[H]
\centering
\caption{\textbf{Out-of distribution prediction:} Mean and standard deviation of relative $L_2$ error for the predicted velocity field of different velocity models predicted using DeepONet and InversionNet. The DeepONet and InversionNet show a similar accuracy except CurveVel-B. In the case of CurveVel-B, the mean and standard deviation for InversionNet are higher than that of DeepONet.}
\label{Table:Out of distribution}
\begin{tabular}{L{2.5cm}|ccccc} \hline
\multicolumn{1}{c|}{Method}  & FlatFault-B & CurveVel-A & CurveVel-B & CurveFault-A & CurveFault-B \\ \hline
DeepONet & (14.66, 3.65) & (12.97, 5.16)  & (30.01, 11.32) & (11.54, 5.47)  & (18.99, 3.82) \\
InversionNet & (13.60, 3.95) & (11.44, 5.29) & (37.73, 14.88) & (9.16, 4.96)  & (18.39, 4.10) \\ \hline
\end{tabular}
\end{table}
\subsection{Application of DeepONet to Create a Starting Model for FWI}
\label{Subsection:Results:Lagecy FWI}
\begin{figure}[H]
    \includegraphics[width=1\textwidth]{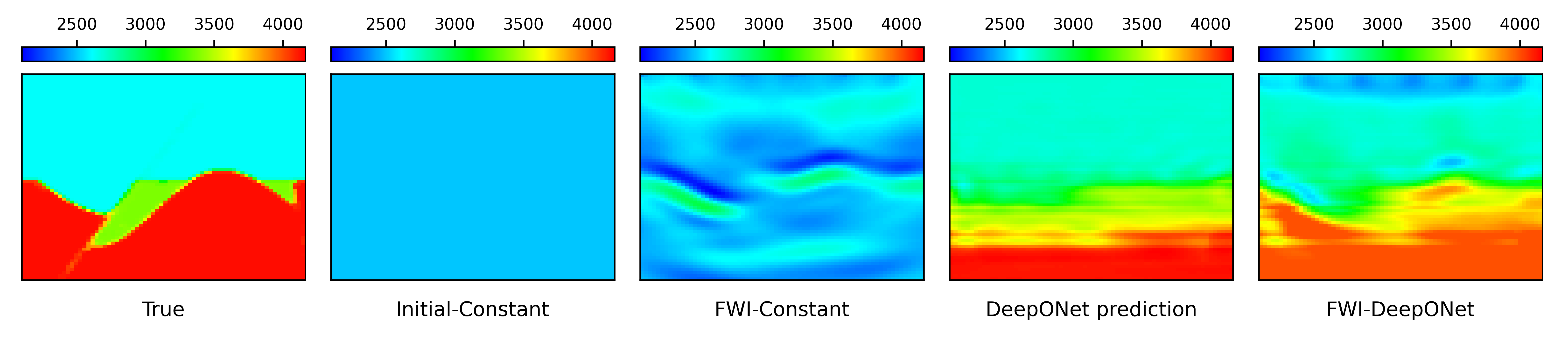}
    \caption{\textbf{Hybrid model:} The predicted velocity field using the proposed hybrid framework. In the first column, we have shown the true velocity model. The second column is a homogeneous initial velocity model, and the third column is the velocity field from applying FWI with this homogeneous starting model. The fourth column is the predicted velocity field from DeepONet and is considered for the hybrid model to perform FWI. In the fifth column, we have shown the results obtained from FWI considering the DeepONet output as the starting velocity model. Not surprisingly, FWI performed using the DeepONet result as a starting model shows better accuracy.}
    \label{fig_legacy_fwi}
\end{figure}

To tackle the challenge of selecting a starting velocity model in an FWI workflow, we conducted two computational experiments. In the first experiment, we created a starting velocity model using the inference from DeepONet and performed FWI on the data using the methods proposed by \citeauthor{Virieux_2009} \cite{Virieux_2009}. We compared this with using a constant starting velocity model as the starting point for the FWI workflow. The results of this exercise are shown in \Cref{fig_legacy_fwi}. Not surprisingly, it is clear that the velocity model obtained using the DeepONet-based starting velocity model is more accurate compared to using a constant starting velocity model. Future tests should quantify how this compares with starting velocity models constructed with more traditional methods such as travel-time inversion. Without such tests against alternative approaches to creating starting models, we cannot provide a stronger statement regarding the utility of the DeepONet result. 
\section{Conclusions}
\label{Section:Summary}
In this study, we have developed a DeepONet-based operator to predict the subsurface velocity field from the seismic waveforms recorded at the surface. The overall  $L_2$ error is slightly higher than that of InversionNet \cite{OpenFWI} when considering clean noise-free seismic waveforms. However, the DeepONet model is less sensitive to noisy data than InversionNet, suggesting that DeepONet is more robust than InversionNet. We propose that the DeepONet results can be used as a starting model  traditional FWI, although we have not tested its performance against other methods of producing starting models such as travel-time analysis. The proposed integration of DeepONet and traditional FWI may accelerate the inversion process and may also enhance reliability and robustness. 
\par The poor accuracy of the DeepONet predictions near discontinuities like faults or interfaces might be improved by considering more points near these regions. This approach would require an unstructured (and non-uniform) mesh, and the element size would be determined by the local velocity present in the model \cite{atmicMeshing}. The resolution of the inverted velocity field near discontinuities may also be improved by considering an additional loss term which focuses on locations where there are sharp changes in velocity. Further study is required in this direction to improve the accuracy of the DeepONet prediction. In addition, this study could potentially be improved by a novel two-step training method \cite{Lee_2024_2step}.  Finally, future studies may also benefit from an efficient transfer learning strategy to generalize the DeepONet model and include physics in the DeepONet training. 
\section*{Acknowledgement}
We acknowledge financial support from Shell USA. This research was conducted using computational resources and services at the Center for Computation and Visualization, Brown University. Additionally, we thank the AFOSR Grant FA9550-23-1-0671 for supporting the purchase of modern GPU hardware. Finally, KS acknowledges support from the Air Force Office of Scientific Research (AFOSR) under MURI grant FA9550-20-1-0358.

\section*{Training-Testing data}
We considered the data generated by \citeauthor{OpenFWI} \cite{OpenFWI} for training and testing of the neural network model in the present study. For comparison of DeepONet results, we considered the InversionNet model trained by \citeauthor{OpenFWI} \cite{OpenFWI}.

\section*{Declaration of competing interest}
The authors declare that they have no known competing financial interests or personal relationships that could have appeared to influence the work reported in this paper.
\bibliography{Reference}
\bibliographystyle{unsrtnat}

\clearpage
\appendix
\renewcommand{\thesection}{\Alph{section}}
\renewcommand{\thesubsection}{\Alph{section}.\arabic{subsection}}
\renewcommand{\thesubsubsection}{\Alph{section}.\arabic{subsection}.\arabic{subsection}}

\setcounter{equation}{0}
\renewcommand{\theequation}{\thesection.\arabic{equation}}
\setcounter{table}{0}
\renewcommand{\thetable}
{\Alph{section}.\arabic{table}}
\setcounter{figure}{0}
\renewcommand{\thefigure}{\Alph{section}.\arabic{figure}}
\section{Neural network size}
The neural network size considered for the proposed DeepONet is shown in \Cref{Table:DeepONet size}.
\begin{table}[H]
\centering
\caption{\textbf{DeepONet size:} DeepONet Network size considered in the present study. In the case of CNN, the size represents as follows: the first two numbers denote the 2D kernel size, and the third and fourth numbers denote the input and output channel size, respectively. Similarly, in the case of TrCNN, the first two numbers denote 2D kernel size, and the third and fourth numbers denote the output and input channel, respectively. Stride is the stride considered in each direction in the convolution process. We consider valid padding in the CNN and TrCNN. The two numbers shown in the bracket in TrCNN and the first layers of CNN-1 indicate one input from the previous layer and one input from the skip connection, as shown in the DeepONet architecture.}
\label{Table:DeepONet size}
\begin{tabular}{C{3cm}|L{3cm}|C{1cm}|C{2cm}}
\hline \hline
\multicolumn{4}{c}{UNet for branch network} \\ \hline
Layers  & Size & Stride & Activation function  \\ \hline
CNN-1 & $5-9-5-8$ & $2-2$ & ReLu \\
CNN-2 & $5-9-8-10$ & $2-2$ & ReLu \\
CNN-3  & $5-9-10-12$ & $2-2$ & ReLu \\
CNN-4  & $5-9-12-16$ & $2-2$ & ReLu \\ \hline
TrCNN-1  & $5-9-8-16$ & $2-2$ & Leaky ReLu \\
TrCNN-2  & $5-9-12-(8+12)$ & $2-2$ & Leaky ReLu \\
TrCNN-3  & $5-9-16-(12+10)$ & $2-2$ & Leaky ReLu \\
TrCNN-4  & $5-9-20-(16+8)$ & $2-2$ & Leaky ReLu \\ \hline \hline
 \multicolumn{4}{c}{CNN for branch network} \\ \hline
 Layers  & Size & Stride & Output size  \\ \hline
CNN-1 & $3-9-(20+5)-32$ & $1-1$ & ReLu \\
CNN-2 & $3-9-32-40$ & $1-1$ & ReLu \\
CNN-3 & $3-9-40-48$ & $1-1$ & ReLu \\
CNN-4 & $3-9-48-64$ & $1-2$ & ReLu \\
CNN-5 & $3-9-64-64$ & $1-2$ & ReLu \\
CNN-6 & $3-9-64-128$ & $2-2$ & ReLu \\
CNN-7 & $3-9-128-128$ & $2-2$ & ReLu \\
CNN-8 & $3-9-128-256$ & $2-2$ & ReLu \\
CNN-9 & $3-9-256-256$ & $2-2$ & ReLu \\ \hline \hline
\multicolumn{4}{c}{DNN for branch network} \\ \hline
  & \multicolumn{2}{c|}{FCN/DNN size}  & Activation function  \\ \hline
FCN/DNN  & \multicolumn{2}{c|}{$4096-2000\times4-2000$} & ReLu \\ \hline

\multicolumn{4}{c}{FCN/DNN for Trunk network} \\ \hline
  & \multicolumn{2}{c|}{FCN/DNN size}  & Output size  \\ \hline
FCN/DNN & \multicolumn{2}{c|}{$2-2000\times10-2000$} & ReLu \\ \hline
 \end{tabular}
\end{table}

\end{document}